\UseRawInputEncoding
\documentclass[11pt]{article}
\usepackage[preprint]{acl}

\usepackage{times}
\usepackage{latexsym}

\usepackage[T1]{fontenc}
\usepackage{inconsolata}

\usepackage[utf8]{inputenc}

\usepackage{microtype}
\usepackage{booktabs}

\usepackage{inconsolata}

\usepackage{graphicx}
\usepackage{subcaption}

\usepackage{listings}
\usepackage{xcolor}

\usepackage{multirow}
\usepackage{colortbl}
\usepackage{array}
\usepackage{tabularx}

\newcolumntype{L}{>{\raggedright\arraybackslash}X}
\newcolumntype{C}{>{\centering\arraybackslash}X}
\newcolumntype{R}{>{\raggedleft\arraybackslash}X}

\definecolor{LightGreen}{RGB}{232,245,233}
\definecolor{LightRed}{RGB}{253,234,234}
\definecolor{HeaderGray}{RGB}{245,247,250}
\definecolor{PromptBlue}{RGB}{237,246,255}
\definecolor{PromptBlueFrame}{RGB}{64,116,170}
\definecolor{PromptPurple}{RGB}{248,244,255}
\definecolor{PromptPurpleFrame}{RGB}{126,87,194}
\definecolor{ArmTwoBg}{RGB}{246,250,255}
\definecolor{ArmThreeBg}{RGB}{251,248,255}
\newcommand{\armtwo}[1]{\cellcolor{ArmTwoBg}#1}
\newcommand{\armthree}[1]{\cellcolor{ArmThreeBg}#1}
\newcolumntype{A}{>{\columncolor{ArmTwoBg}\centering\arraybackslash}c}
\newcolumntype{B}{>{\columncolor{ArmThreeBg}\centering\arraybackslash}c}

\definecolor{TraceRawBg}{RGB}{255,248,248}
\definecolor{TraceRawFrame}{RGB}{177,91,91}
\definecolor{TraceWorkflowBg}{RGB}{255,251,242}
\definecolor{TraceWorkflowFrame}{RGB}{181,133,54}
\definecolor{TraceSkillBg}{RGB}{244,251,246}
\definecolor{TraceSkillFrame}{RGB}{76,137,94}
\definecolor{TraceOutcomeBg}{RGB}{247,248,250}
\definecolor{TraceOutcomeFrame}{RGB}{119,128,140}

\newcommand{\traceheading}[2]{%
  \par\noindent
  \colorbox{#1}{%
    \parbox{\dimexpr\linewidth-2\fboxsep\relax}{%
      \sffamily\bfseries\small\color{white}#2}}
  \par\smallskip
}

\newcommand{\traceoutcome}[1]{%
  \par\noindent
  \fcolorbox{TraceOutcomeFrame}{TraceOutcomeBg}{%
    \parbox{\dimexpr\linewidth-2\fboxrule-2\fboxsep\relax}{%
      \small\textbf{Outcome.} #1}}
  \par\medskip
}

\lstdefinestyle{trace-base}{%
  basicstyle=\ttfamily\scriptsize,
  frame=single,
  framerule=0.45pt,
  framesep=5pt,
  numbers=none,
  xleftmargin=0pt,
  xrightmargin=0pt,
  aboveskip=0pt,
  belowskip=7pt,
  breaklines=true,
  breakatwhitespace=false,
  columns=fullflexible,
  keepspaces=true,
  showstringspaces=false
}
\lstdefinestyle{trace-raw}{style=trace-base,backgroundcolor=\color{TraceRawBg},rulecolor=\color{TraceRawFrame}}
\lstdefinestyle{trace-workflow}{style=trace-base,backgroundcolor=\color{TraceWorkflowBg},rulecolor=\color{TraceWorkflowFrame}}
\lstdefinestyle{trace-skill}{style=trace-base,backgroundcolor=\color{TraceSkillBg},rulecolor=\color{TraceSkillFrame}}

\lstnewenvironment{promptlisting}[2][]%
{\traceheading{PromptBlueFrame}{#2}%
 \lstset{
  basicstyle=\ttfamily\scriptsize,
  breaklines=true,
  breakatwhitespace=false,
  columns=fullflexible,
  keepspaces=true,
  showstringspaces=false,
  frame=single,
  framerule=0.5pt,
  framesep=5pt,
  xleftmargin=0pt,
  xrightmargin=0pt,
  aboveskip=6pt,
  belowskip=10pt,
  rulecolor=\color{PromptBlueFrame},
  backgroundcolor=\color{PromptBlue},
  #1
}}{}

\lstnewenvironment{promptlistingpurple}[1]%
{\traceheading{PromptPurpleFrame}{#1}%
 \lstset{
  basicstyle=\ttfamily\scriptsize,
  breaklines=true,
  breakatwhitespace=false,
  columns=fullflexible,
  keepspaces=true,
  showstringspaces=false,
  frame=single,
  framerule=0.5pt,
  framesep=5pt,
  xleftmargin=0pt,
  xrightmargin=0pt,
  aboveskip=0pt,
  belowskip=10pt,
  rulecolor=\color{PromptPurpleFrame},
  backgroundcolor=\color{PromptPurple}
 }}{}

    \title{Demystifying Agent Skills: Why They Work—Until They Don't\thanks{Preprint.}}

\author{
\begin{tabular}{@{}c@{\quad}c@{\quad}c@{\quad}c@{\quad}c@{}}
\textbf{Zhiyuan Jiang}$^{1,*,\ddagger}$ &
\textbf{Fangrui Huang}$^{3,*}$ &
\textbf{Hanwen Xing}$^{4}$ &
\textbf{Xander Wu}$^{3}$ &
\textbf{Yipeng Gao}$^{4}$ 
\end{tabular}
\\[4pt]
\begin{tabular}{@{}c@{\quad}c@{\quad}c@{\quad}c@{}}
\textbf{Rui Cao}$^{5}$ &
\textbf{Mengdi Wang}$^{1,\dagger}$ &
\textbf{Shilong Liu}$^{1,\dagger}$ &
\textbf{Yijiang Li}$^{2,\dagger}$
\end{tabular}
\\[5pt]
$^{1}$Princeton University \quad
$^{2}$UC San Diego\quad
$^{3}$Stanford University \quad \\
$^{4}$University of Southern California\quad
$^{5}$Johns Hopkins University
\\[3pt]
\small{$^*$Equal contribution; \quad
$^\dagger$Corresponding authors.\quad
$^\ddagger$Work done during an internship at Princeton University.}
}

\begin{document}
\maketitle
\begin{abstract}

Skills have emerged as a practical and effective approach for enhancing LLM agents at inference time through structured packages of knowledge. However, existing evaluations largely measure whether skills improve aggregated task success, leaving a more fundamental question underexplored: \emph{\textbf{When do skills help, why do they work, and where do they fail?}} Through controlled experiments across various benchmarks, agent harnesses and LLMs, we isolate the effects of representation, outcome annotation, retrieval difficulty, and cross-framework robustness of skills. To further answer this question, we design a contrastive study that combines controlled quantitative experiments with paired trajectory analysis. We normalize 8,135 trial records from controlled experiments and retain 238 valid unique labels from 240 open-coded records.
We consolidate these observations into a taxonomy of three high-level categories and twelve skill-use modes: skills work when noisy trajectories become procedural anchors that stabilize execution. Skills improve over Workflow Memory by 6.06 points in matched comparisons. Procedural anchoring accounts for 65.7\% of skill cases, versus 4.5\% for explicit knowledge injection, showing that skills stabilize action rather than inject missing facts. Retrieval is a separate bottleneck: as pools grow from 5 to 100, actual-use precision falls from 29.6\% to 3.3\%. Confusable distractors impair offline identification, yet downstream success remains stable; exact ground-truth invocation is neither sufficient nor necessary. Skills fail under brittle assumptions, incompatible contexts, or insufficient adaptation. These findings move evaluation beyond aggregate success rates and guide reliable self-evolving agents.

\end{abstract}




\begin{figure*}[t]
\centering
\captionsetup[subfigure]{font=small,labelfont=bf,textfont=normalfont,skip=2pt}
\begin{subfigure}[t]{\textwidth}
    \centering
    \includegraphics[width=\linewidth]{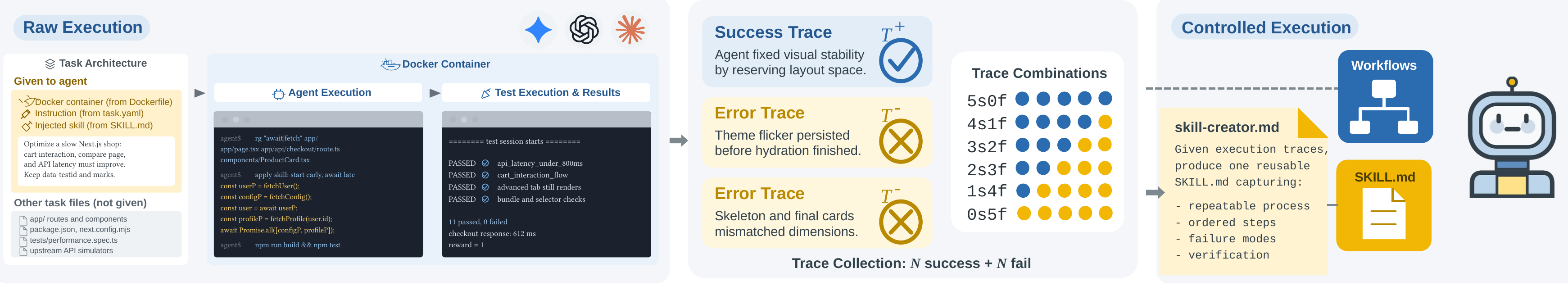}
    \caption{Skill-vs-procedural-memory pipeline.}
    \label{fig:pipeline-procmem-skills}
\end{subfigure}

\begin{subfigure}[t]{\textwidth}
    \centering
    \includegraphics[width=\linewidth]{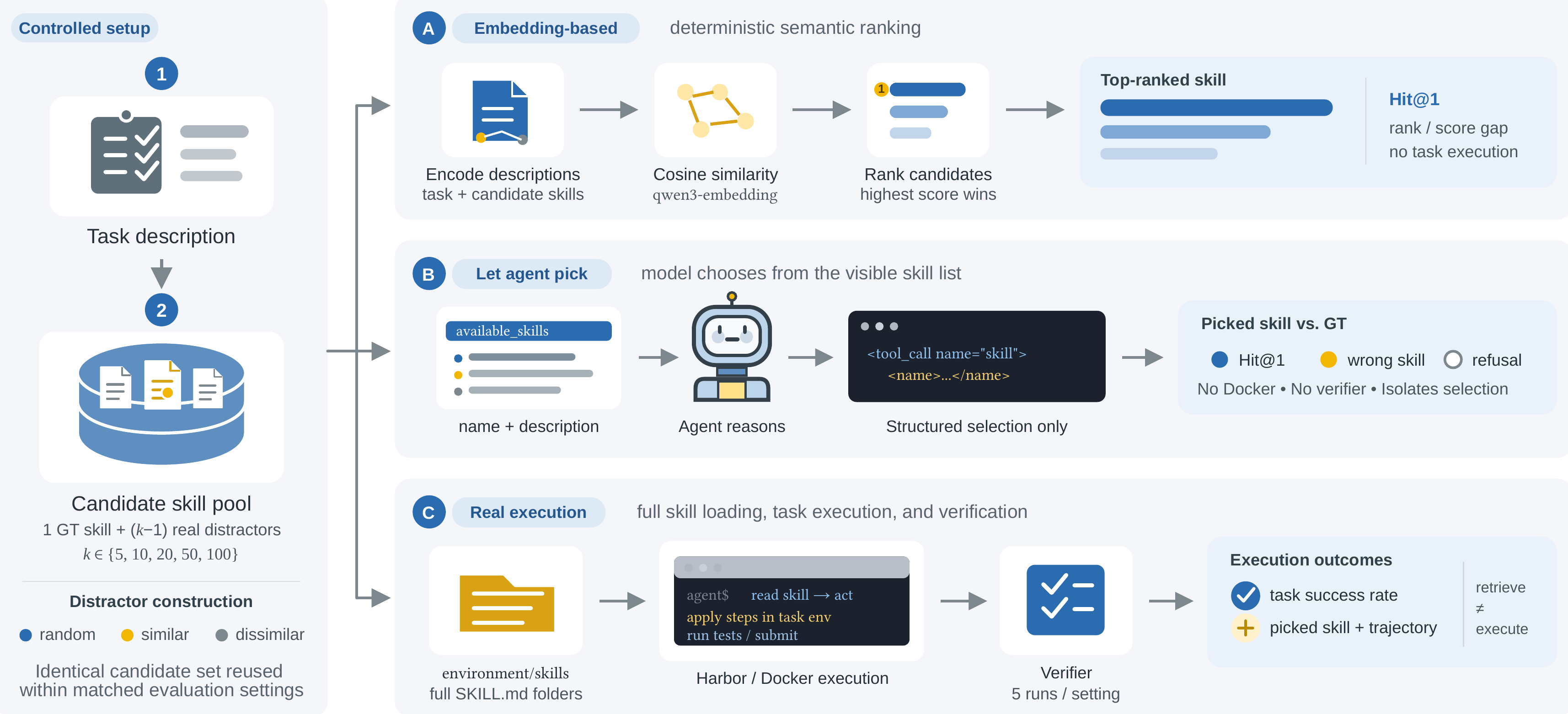}
    \caption{Three-experiment skill-retrieval evaluation.}
    \label{fig:pipeline-retrieval}
\end{subfigure}
\caption{\textbf{Experimental pipelines.} Top: skill versus procedural memory. We execute each task in a fixed Docker environment, collect successful and failed trajectories, and form a fixed-budget composition grid. The same trace pool is distilled either into Workflow Memory or into a reusable \texttt{SKILL.md}, which are then evaluated on matched tasks under the same protocol. Bottom: skill retrieval. Each task is paired with a candidate pool containing its ground-truth skill and $k-1$ real distractors (random, similar, or dissimilar). Matched pools are evaluated independently in three procedures: (A) embedding-based ranking without task execution, (B) explicit agent selection without Docker execution or verification, and (C) full-pool real execution with skill-use parsing after task verification. Outputs from (A) and (B) are not passed to (C).}
\vspace{-4mm}
\label{fig:experimental-pipelines}
\end{figure*}

\begin{figure*}[t]
\centering
\includegraphics[width=\textwidth]{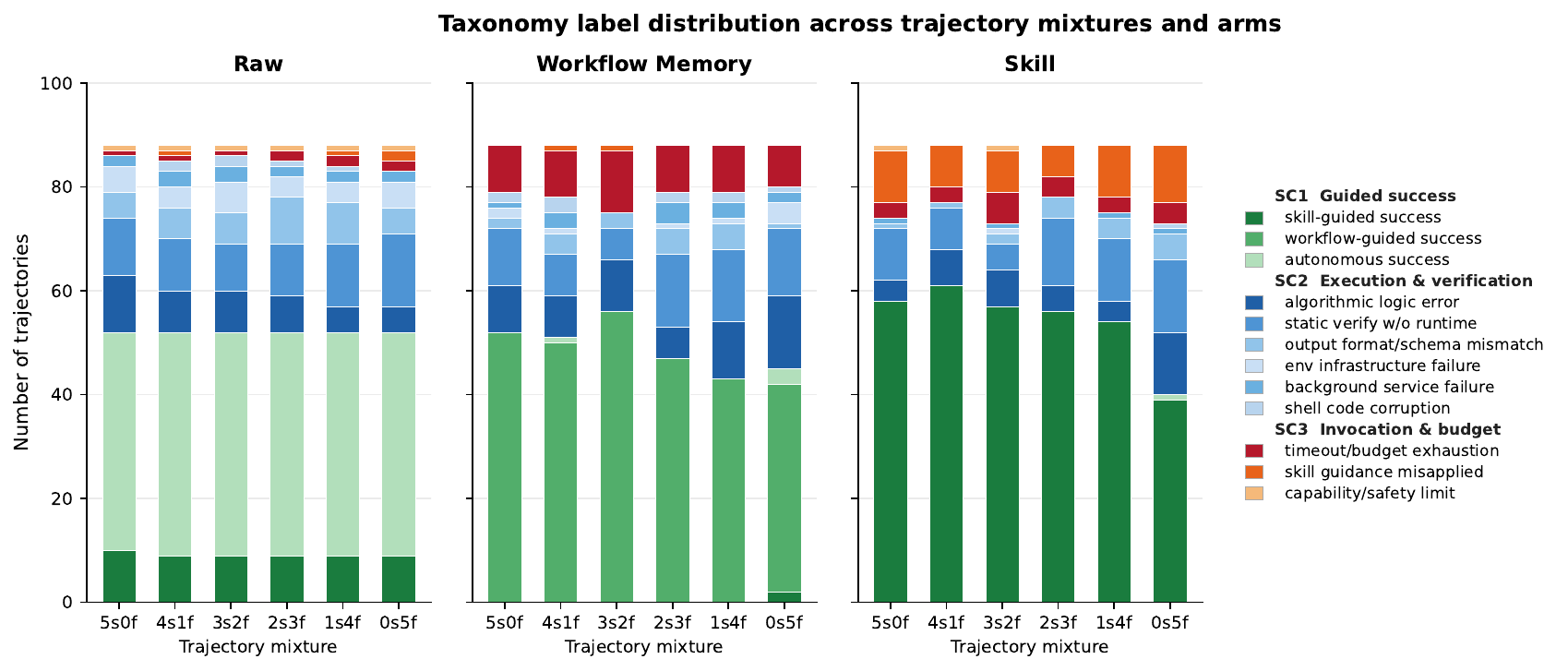}
\caption{\textbf{Taxonomy label distribution across trajectory mixtures and experimental arms.} Stacked bars show trajectory-level labels for Raw, Workflow Memory, and Skill across the six source-trajectory mixtures from $5s0f$ to $0s5f$. Labels are grouped into three high-level categories; per-mode percentages are reported in Appendix Table~\ref{tab:taxonomy-mode-percent}.}
\vspace{-4mm}
\label{fig:taxonomy-per-setting}
\end{figure*}

\section{Introduction}

Large language model (LLM) agents are increasingly expected to improve through experience rather than solve each task from scratch. Recent agent systems therefore store and reuse traces of prior execution: environment setup sequences, tool-use patterns, debugging routines, and verification steps that were discovered in earlier runs. This shift is especially appealing for tool-using agents~\cite{yao2023react,schick2023toolformer,qin2024toolllm,wang2024executable,yang2024sweagent}, where repeated failures often arise not from a lack of high-level reasoning, but from rediscovering the same procedural details again and again.

Among the proposed memory forms, \emph{skills}\cite{anthropic_agent_skills} have emerged as a particularly compelling abstraction. A skill is not simply a record of past execution, but a compact description of what to do, what to check, and what pitfalls to avoid. Compared to storing raw execution traces or direct workflow memories, skills promise three advantages. They can compress noisy experience into a shorter context, standardize procedural knowledge into a stable format, and potentially transfer that knowledge across related tasks.

Despite the significance of skills, existing work studies skills only through aggregated task success: if a skill-augmented agent solves more tasks, the skill is considered useful. Although such evaluations establish that skills do matter, they reveal little about \emph{why} they matter. In particular, they do not explain what fundamentally changes in an agent's behavior before and after a skill is loaded, which parts of execution are stabilized by skill guidance, or why the same skill can help one task while harming another. This leaves the field with a largely empirical view of skill design and improvement: skills are often written, retrieved, and revised through heuristic iteration, prompt tuning, or benchmark-specific trial and error, rather than through a principled understanding of what makes a skill genuinely valuable.

We therefore ask \emph{when skills help, why they work, and where they fail}, moving beyond the preliminary question of \emph{whether skills work}.
Our key contribution is a taxonomy of skill utility and failure, together with a paired research methodology for making skill effects observable rather than treating them as a black box. At a high level, our method compares matched executions with and without skill access and analyzes where the two runs diverge across the skill-use pipeline, including how prior experience is represented, transferred across agent frameworks, retrieved, and invoked. This contrastive view lets us attribute gains and failures to concrete mechanisms instead of aggregate outcomes alone, and provides a more rigorous foundation for studying skill use and for eventually automating the generation of reliably useful skills. 

This perspective suggests that skill utility cannot be understood from end-to-end success alone, but must be analyzed along the stages of the skill-use pipeline. Concretely, we organize our study around four questions: (1) whether representing the same prior experience as a standardized skill differs from injecting it as direct procedural memory; (2) whether skill gains come from the underlying experience itself or from explicit success/failure annotations; (3) whether distilled procedural guidance remains useful when transferred across agent frameworks; and (4) how skill-pool size and confusability affect retrieval and downstream execution.

Through comprehensive controlled experiments and comparative trajectory analysis, we find that skills are most useful when they convert noisy prior experience into compact procedural guidance. The dominant mechanism is not the injection of factual knowledge (4.5\%), but \emph{procedural anchoring} (65.7\%): skills help agents follow more reliable setup steps, tool sequences, implementation routines and verification checks. Consistent with this view, skills reduce several execution-layer failures, including environment setup errors, output-format mismatches, service-lifecycle failures, and shell-command corruption. In comparison, direct workflow memory exposes a complementary weakness: it preserves useful procedural evidence but carries irrelevant exploration, failed branches, and verbose process noise that can increase timeout and drift.

However, our results also show that skills are not universally beneficial. A skill must still be retrieved, interpreted, and applied in the right context. Our retrieval experiments show that retrieval quality cannot be reduced to top-level selection accuracy alone. Hard negative distractors substantially degrade explicit skill selection, but downstream execution is not determined by retrieval accuracy in a one-to-one manner: selecting the correct skill does not guarantee task success, while related non-ground-truth skills can still provide useful procedural support. This reveals a second failure boundary of skills: even when a useful skill exists in the pool, the agent may fail because the skill is retrieved for the wrong procedural context, invoked only superficially, or insufficient to overcome task-level execution bottlenecks such as timeout, numerical precision, missing dependencies, and brittle implementation requirements.

These findings motivate a lifecycle view of skill use. Skills work when prior experience is distilled into reusable procedural anchors that are retrieved and invoked in compatible contexts. They fail when the distilled guidance is noisy, over-specific, mismatched to the current task, or followed without adaptation. Thus, skill-based self-improvement is not merely a problem of accumulating more memories, but of building agents that can generate, retrieve, and apply procedural abstractions reliably.

Overall, our contributions can be summarized as follows:
\begin{itemize}
    \item We formulate a systematic analysis of \emph{when skills help, why they work, and where they fail}, shifting skill evaluation from aggregate success rates to the mechanisms by which skills change agent behavior.

    \item We introduce a contrastive trajectory-analysis methodology for studying skill effects. We normalize 8,135 trial records, perform open coding over 240 sampled trajectories, and consolidate 238 valid unique labels into a taxonomy of three high-level categories and twelve skill-use modes.

    \item We conduct controlled skill-vs-workflow experiments that isolate how the same prior experience behaves when represented as direct procedural memory or as a distilled skill. Our results show that skills primarily work as procedural anchors, while workflow memory often preserves verbose exploration, failed branches, and process noise.

    \item We analyze retrieval and downstream execution jointly, showing that retrieval accuracy alone does not explain task success. Skills fail not only when retrieval misses the right artifact, but also when retrieved guidance is procedurally incompatible, weakly invoked, or insufficient for the execution bottleneck.

\end{itemize}



\begin{table*}[t]
\centering
\footnotesize
\setlength{\tabcolsep}{3.5pt}
\renewcommand{\arraystretch}{1.08}
\resizebox{\textwidth}{!}{%
\begin{tabular}{ll|*{2}{>{\centering\arraybackslash}m{1.35cm}}|*{2}{>{\centering\arraybackslash}m{1.35cm}}|*{2}{>{\centering\arraybackslash}m{1.35cm}}}
\toprule
\multirow{2}{*}{\textbf{Agent + Model}} &
\multirow{2}{*}{\textbf{Trajectory Mix}} &
\multicolumn{2}{c|}{\textbf{Terminal-Bench-2}} &
\multicolumn{2}{c|}{\textbf{SkillsBench}} &
\multicolumn{2}{c}{\textbf{Terminal-Bench-Pro}} \\
& & \textbf{Workflow} & \textbf{Skill} & \textbf{Workflow} & \textbf{Skill} & \textbf{Workflow} & \textbf{Skill} \\
\midrule
\rowcolor{HeaderGray}
\multicolumn{2}{l|}{\textit{Codex + GPT-5.3-Codex Raw}} &
\multicolumn{2}{c|}{0.5935} & \multicolumn{2}{c|}{0.5083} & \multicolumn{2}{c}{0.5394} \\
\multirow{6}{*}{\shortstack[l]{Codex\\GPT-5.3-Codex}}
& 5s0f & \cellcolor{LightRed}\textbf{0.4452} & \cellcolor{LightGreen}0.7548 & \cellcolor{LightGreen}0.5250 & \cellcolor{LightGreen}\textbf{0.7250} & \cellcolor{LightGreen}\textbf{0.7333} & \cellcolor{LightGreen}0.7455 \\
& 4s1f & \cellcolor{LightRed}0.4000 & \cellcolor{LightGreen}0.7290 & \cellcolor{LightGreen}0.5667 & \cellcolor{LightGreen}0.6167 & \cellcolor{LightGreen}\textbf{0.7333} & \cellcolor{LightGreen}\textbf{0.7939} \\
& 3s2f & \cellcolor{LightRed}0.4194 & \cellcolor{LightGreen}\textbf{0.7806} & \cellcolor{LightGreen}\textbf{0.6417} & \cellcolor{LightGreen}0.6250 & \cellcolor{LightGreen}0.6970 & \cellcolor{LightGreen}0.7333 \\
& 2s3f & \cellcolor{LightRed}0.3677 & \cellcolor{LightGreen}0.6839 & \cellcolor{LightGreen}0.6083 & \cellcolor{LightGreen}0.7083 & \cellcolor{LightGreen}0.6667 & \cellcolor{LightGreen}0.6667 \\
& 1s4f & \cellcolor{LightRed}0.2710 & \cellcolor{LightGreen}0.7097 & \cellcolor{LightGreen}0.5167 & \cellcolor{LightGreen}0.6167 & \cellcolor{LightGreen}0.6121 & \cellcolor{LightGreen}0.5818 \\
& 0s5f & \cellcolor{LightRed}0.2839 & \cellcolor{LightRed}0.5161 & \cellcolor{LightGreen}0.5833 & \cellcolor{LightRed}0.4500 & \cellcolor{LightRed}0.4788 & \cellcolor{LightRed}0.4303 \\
\midrule
\rowcolor{HeaderGray}
\multicolumn{2}{l|}{\textit{Gemini CLI + Gemini-3.1-Pro-Preview Raw}} &
\multicolumn{2}{c|}{0.5000} & \multicolumn{2}{c|}{0.4762} & \multicolumn{2}{c}{0.5615} \\
\multirow{6}{*}{\shortstack[l]{Gemini CLI\\Gemini-3.1-Pro-Preview}}
& 5s0f & \cellcolor{LightGreen}0.6231 & \cellcolor{LightGreen}\textbf{0.7923} & \cellcolor{LightGreen}0.5524 & \cellcolor{LightGreen}\textbf{0.7429} & \cellcolor{LightRed}0.5308 & \cellcolor{LightGreen}\textbf{0.6692} \\
& 4s1f & \cellcolor{LightGreen}0.5308 & \cellcolor{LightGreen}0.7615 & \cellcolor{LightGreen}0.5238 & \cellcolor{LightGreen}0.6190 & \cellcolor{LightGreen}\textbf{0.6923} & \cellcolor{LightGreen}0.6308 \\
& 3s2f & \cellcolor{LightGreen}\textbf{0.6462} & \cellcolor{LightGreen}0.7462 & \cellcolor{LightGreen}\textbf{0.5619} & \cellcolor{LightGreen}0.6667 & \cellcolor{LightGreen}0.6462 & \cellcolor{LightRed}0.5462 \\
& 2s3f & \cellcolor{LightGreen}0.6000 & \cellcolor{LightGreen}0.7000 & \cellcolor{LightGreen}0.5524 & \cellcolor{LightGreen}0.6762 & \cellcolor{LightGreen}0.6846 & \cellcolor{LightRed}0.5077 \\
& 1s4f & \cellcolor{LightGreen}0.5846 & \cellcolor{LightGreen}0.6923 & \cellcolor{LightGreen}0.4857 & \cellcolor{LightGreen}0.6000 & \cellcolor{LightGreen}0.5923 & \cellcolor{LightGreen}0.5692 \\
& 0s5f & \cellcolor{LightGreen}0.5231 & \cellcolor{LightRed}0.4769 & \cellcolor{LightRed}0.4286 & \cellcolor{LightRed}0.4095 & \cellcolor{LightRed}0.4769 & \cellcolor{LightRed}0.4615 \\
\bottomrule
\end{tabular}}

\caption{\textbf{Task success rates for Workflow Memory and Skill injection across trajectory mixtures.} Gray rows denote Raw baselines. Green, red, and unshaded cells indicate values above, below, and equal to the corresponding Raw baseline, respectively; bold marks the row-wise maximum across mixture settings. Mixture labels denote the numbers of successful (s) and failed (f) source trajectories. Terminal-Bench-Pro rates use 130 trials per condition, with infrastructure or verifier errors counted as failures.}
\label{tab:main-workflow-skill-results}
\end{table*}

\section{Related Works}

\subsection{Memory Reuse in LLM Agents}

Recent work has studied memory as a mechanism for enabling LLM agents to adapt across tasks and interactions. Early systems store and retrieve episodic experiences, reflections, or interaction histories to support future planning and decision-making~\citep{park2023generativeagentsinteractivesimulacra,shinn2023reflexionlanguageagentsverbal,wang2023voyageropenendedembodiedagent,zhong2023memorybankenhancinglargelanguage}. This direction has expanded to structured external memory~\citep{packer2024memgptllmsoperatingsystems,xu2025amemagenticmemoryllm}, retrieval-augmented experience reuse~\citep{zhao2024expelllmagentsexperiential,zhou2025memento,zhang2026agenticcontextengineering,wu2026proceduralknowledgescale}, procedural-memory management~\citep{belikova2026managingproceduralmemory}, and task-agnostic memory graphs that compress episodic observations into knowledge-centric structures~\citep{yang2026plugmemtaskagnosticpluginmemory}. Recent surveys frame agent memory as a write-manage-read loop and identify persistent challenges in consolidation, retrieval, contradiction handling, and evaluation~\citep{zhang2024surveymemorymechanismlarge,du2026memoryautonomousllmagentsmechanisms}.

A related line of work shifts from remembering \emph{what happened} to reusing \emph{how to act}. Workflow memories summarize past trajectories into action-level procedures for future tasks~\citep{wang2024agentworkflowmemory,fang2026mempexploringagentprocedural}. Skill-based systems make such procedural knowledge more explicit by storing reusable procedures as first-class artifacts, including executable routines, scripts, skill directories, or hierarchical skill knowledge bases~\citep{ni2026trace2skilldistilltrajectorylocallessons,yang2026autoskill,zhang2026memskill,wang2026skillx,jiang2026sokagenticskills}. Recent work further studies skill activation and selection during execution~\citep{xia2026skillrlevolvingagentsrecursive,zheng2026skillrouterskillroutingllm}, as well as automatic skill discovery, revision, and evolution through experience~\citep{mi2026skillprolearningreusableskills,alzubi2026evoskillautomatedskilldiscovery,lei2026skillevolbench,lin2026museautoskill,he2026reskill}. However, skills are still often evaluated through aggregate task success, obscuring how they differ from broader procedural memory and why they help or fail in specific executions. We address this gap by comparing skills against procedural memory and attributing their effects to stages of representation, retrieval, invocation, transfer, and abstraction.

\subsection{Benchmarks for Evaluating Agent Behavior}

Existing agent benchmarks provide natural settings for studying agent behavior. General tool-use and computer-use benchmarks evaluate multi-step decision-making, tool invocation, and environment interaction \citep{wang2024officebenchbenchmarkinglanguageagents,xu2023toolmanipulationcapabilityopensource,mialon2023gaiabenchmarkgeneralai,deng2023mind2webgeneralistagentweb,zhou2023webarena,koh2024visualwebarena,drouin2024workarena,xu2024theagentcompany,xie2024osworld,rawles2025androidworld,trivedi2024appworld,chen2026appworldul,yao2024taubench,barres2025tau2bench}. Software and terminal benchmarks\cite{jimenez2024swebenchlanguagemodelsresolve,wang2025swebenchframeworkscalablegeneration,merrill2026terminalbenchbenchmarkingagentshard,yang2024sweagent,wang2024executable} are particularly relevant because they require long-horizon execution over files, tests, and environment states. Especially, Terminal-Bench\cite{merrill2026terminalbenchbenchmarkingagentshard} provides realistic command-line tasks with isolated environments and test-based verification, making it well suited for observing whether procedural guidance improves execution.

Recent work further introduces benchmarks that explicitly test agent skills. SkillsBench\cite{li2026skillsbenchbenchmarkingagentskills} measures structured skills across diverse tasks, while SWE-Skills-Bench\cite{han2026sweskillsbenchagentskillsactually} studies the marginal utility of skill documents in real-world software-engineering settings. However, these benchmarks mainly report final task success. Our work extends this analysis perspective to skill-based memory by connecting success and failure modes to specific memory decisions: which skill is retrieved, how it is interpreted, and whether its procedural abstraction matches the current task.

\section{Study Design}
\subsection{Research Questions}

\emph{When do skills help, why do they work, and where do they fail?} Rather than treating skill use as a black-box performance improvement, we analyze it as a controlled transformation of prior agent experience into procedural knowledge.

\textbf{RQ1: How does procedural representation shape experience reuse?}
Prior executions can be reused either as direct workflow memories or as distilled skills. Both representations expose the agent to procedural experience, but they differ in how that experience is packaged: workflow memory preserves trace-level execution details, while a skill compresses them into a standardized procedural artifact. RQ1 asks how this representational form affects downstream behavior when the underlying trajectories are held fixed. In particular, we compare whether workflow memory and skills both act as procedural anchors, and whether skills provide additional robustness by reducing trace dependence, context burden, or framework-specific coupling.

\textbf{RQ2: What role do outcome signals play in learning from trajectories?}
Prior trajectories may help because they contain reusable procedures, or because outcome information indicates which behaviors succeeded or failed. RQ2 isolates the role of these signals by comparing standard settings with no-hint settings, where the same trajectories are used but explicit success/failure annotations are removed. This tests whether the benefit of prior experience comes mainly from procedural content itself or from outcome labels that guide selection, distillation, and use.

\textbf{RQ3: Does distilled procedural guidance transfer across agent frameworks?}
Procedural knowledge may be coupled to the prompting style, tool interface, and execution loop of the framework that produced it. RQ3 studies whether workflow memories and skills constructed from trajectories in one agent framework remain useful when evaluated in another. By holding the source experience fixed and changing the target framework, we test whether distillation into skills provides a more portable representation than direct workflow memory.

\textbf{RQ4: How does skill-pool construction affect retrieval and downstream use?}
In realistic settings, skills are selected from a library rather than provided directly. RQ4 examines how retrieval quality changes as the skill pool becomes larger or more confusable. We evaluate retrieval both as an isolated skill-selection problem and in downstream execution, varying pool size and distractor type. This lets us distinguish failures caused by not finding the right skill from failures that occur after a skill is available but is applied, adapted, or verified incorrectly.

\subsection{Experimental Setup}

\paragraph{Model and benchmark selection.}
We evaluate skill use under complementary controlled settings. For the skill-vs-procedural-memory and no-hint studies in RQ1--RQ2, we instantiate the same experimental protocol with two agent--model pairings: Codex + GPT-5.3-Codex and Gemini CLI + Gemini-3.1-Pro-Preview. RQ3 constructs prior-experience artifacts from the primary Codex setting and evaluates their cross-framework transfer in Gemini CLI with Gemini-3.1-Pro-Preview. RQ4 uses Qwen3-Embedding-0.6B~\citep{zhang2025qwen3embedding} for embedding retrieval, and evaluates explicit selection and downstream execution with Gemini CLI + Gemini-3.1-Pro-Preview and Codex + GPT-5.4. GPT-5.3-Codex was no longer available under the same evaluation access when RQ4 was conducted, so GPT-5.4 was used for the Codex pairing in RQ4. Accordingly, RQ4 is interpreted strictly as a within-pairing comparison among its three independent experiments; its absolute values are not directly compared with the RQ1--RQ3 results, whose Codex experiments use GPT-5.3-Codex. Our evaluation suite combines Terminal-Bench~\cite{merrill2026terminalbenchbenchmarkingagentshard} and SkillsBench~\cite{li2026skillsbenchbenchmarkingagentskills}. Downstream execution experiments follow Harbor's standard evaluation workflow~\cite{Harbor_Framework_Team_Harbor_A_framework_2026} with $n=5$ unique trials per task and a parallelism of $20$ unless otherwise noted; retrieval-isolation experiments use one query per task--pool setting. In skill-based conditions, skills are placed in the execution environment as reusable procedural resources rather than fully incorporated into the initial context~\citep{anthropic_agent_skills}. The complete agent, model, benchmark and task-split information is provided in Appendix~\ref{model_details}.

\subsection{Experimental Design}


To answer these questions, we design a set of controlled studies, as well as fine-grained trajectory analyses that examine the full skill-use pipeline. Full implementation details can be found in Appendix\ref{app:experimental-details}.

\paragraph{Controlled study of procedural experience representation (RQ1--RQ2).}
We first study how prior agent experience becomes reusable procedural knowledge. The key question is whether skills help simply because they expose the agent to past trajectories, or because distilling those trajectories into a standardized procedural form changes how the agent can use them. To isolate this effect, we compare three conditions: \emph{Raw}, which receives no prior experience; \emph{Workflow Memory}, which receives cleaned procedural traces from prior executions; and \emph{Skill}, which receives a standardized \texttt{SKILL.md} distilled from the same workflows.

For each selected task, we collect successful and failed raw trajectories and construct a balanced trajectory pool. We then instantiate a fixed-budget composition grid, varying the source evidence from success-only to failure-only, e.g., $5s0f$ through $0s5f$. For every composition, Workflow Memory and Skill are built from the same selected trajectories and evaluated on the same target tasks. This holds the underlying experience constant while varying only its representation. To separate procedural content from explicit outcome signals during skill construction, we further create \emph{standard} and \emph{no-hint} Skill variants. In the standard setting, success/failure identities are visible to the skill creator; in the no-hint setting, these annotations are removed while preserving the same trajectories and execution protocol.

\paragraph{Cross-framework transfer evaluation (RQ3).}
RQ3 tests whether procedural knowledge remains useful when moved outside the agent framework in which it was produced. We construct skills and workflow memories from trajectories collected in the primary Codex setting, then evaluate them in Gemini CLI with Gemini-3.1-Pro-Preview. This setting isolates framework transfer: the source experience is fixed, but the target agent differs in prompting style, tool-use interface, and execution behavior. Comparing transferred Skill and Workflow Memory against the target-framework Raw baseline lets us examine whether distilled skills preserve reusable procedural guidance more robustly than direct workflow traces.

\paragraph{Controlled study of skill retrieval and downstream execution (RQ4).}
RQ4 first evaluates whether a growing and increasingly confusable skill library remains operationally usable during task execution. We make the complete candidate pool available to the agent, without preselecting skills, and measure both the ground-truth overlap of the skills accessed during execution and the final verifier outcome. To characterize whether the relevant skills are also difficult to identify outside the execution setting, we conduct two additional offline diagnostics on matched candidate pools. An embedding retriever ranks skills using task--description similarity, while an agent explicitly selects potentially useful skills without executing the task. These diagnostics are run independently, and their outputs are not passed to the execution experiment. Because the Codex pairing in this study uses GPT-5.4 for availability reasons, all RQ4 results are interpreted as within-pairing comparisons among these three experiments and should not be interpreted as direct model-to-model or cross-RQ comparisons with the GPT-5.3-Codex results reported for RQ1--RQ3.

Throughout, these experiments are denoted Arm 1 (embedding retrieval), Arm 2 (explicit agent selection), and Arm 3 (full-pool real execution), respectively.

All three experiments use the same controlled candidate-pool construction. Each pool contains the task's ground-truth skill set and distractors; pool size ranges from 5 to 100, and distractors are sampled as random, semantically similar, or dissimilar skills~\citep{reimers2019sentencebert,thakur2021beir,muennighoff2023mteb}. The resulting measurements characterize complementary aspects of skill use: offline identification quality, execution-time skill access, and downstream task completion. They are compared as independent measurements rather than as sequential stages, and no selection output is transferred from either offline diagnostic to the execution experiment. Full implementation details can be found in Appendix~\ref{rq4}.

\begin{figure*}[t]
\centering
\begin{subfigure}[t]{0.46\textwidth}
\centering
\includegraphics[width=\linewidth]{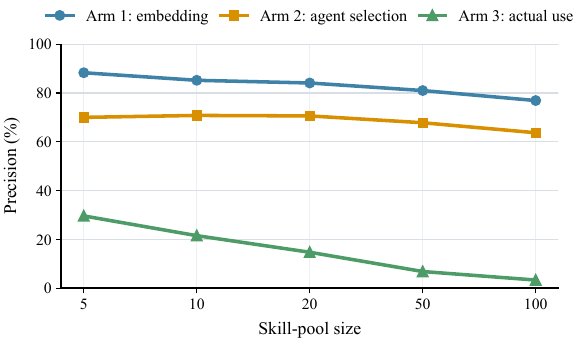}
\caption{Precision across the three independent RQ4 experiments.}
\label{fig:retrieval-pipeline-precision}
\end{subfigure}\hfill%
\begin{subfigure}[t]{0.52\textwidth}
\centering
\includegraphics[width=\linewidth]{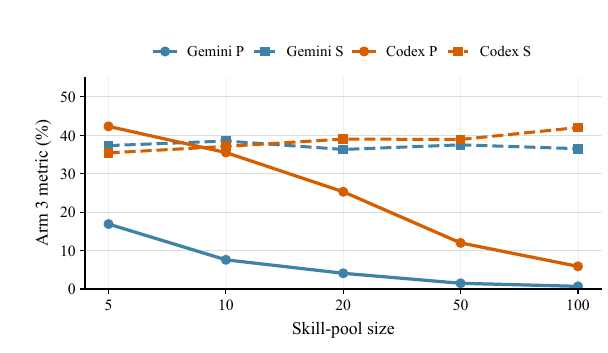}
\caption{Arm 3 skill-use precision and downstream success.}
\label{fig:retrieval-arm3-precision-success}
\end{subfigure}
\caption{\textbf{Skill retrieval and execution-time skill use on SkillsBench.} Left: precision for the two offline diagnostics. Right: parsed actual-use precision (solid lines) and downstream success (dashed lines) for Arm 3. Curves average over random, similar, and dissimilar pool regimes; outputs are not passed between experiments.}
\label{fig:retrieval-combined}
\end{figure*}

\begin{figure}[t]
\centering
\includegraphics[width=\linewidth]{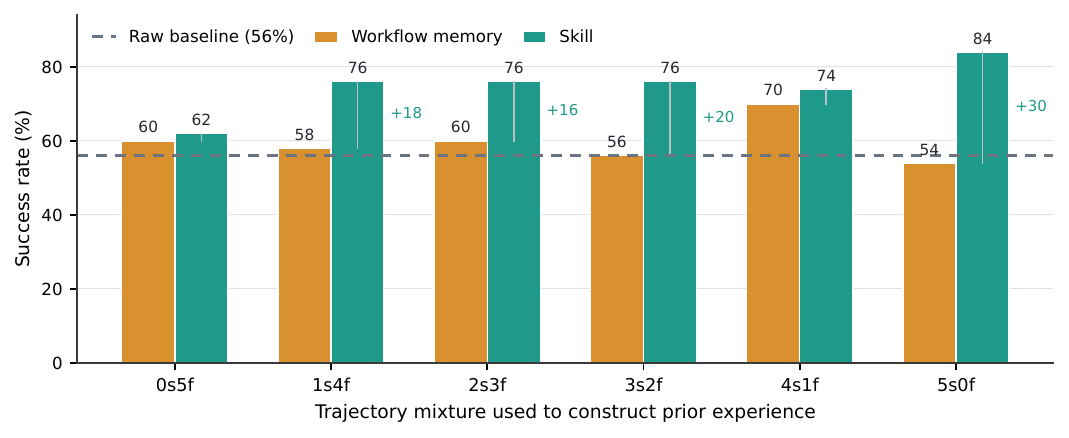}
\caption{\textbf{Cross-framework transfer of procedural experience.} Prior-experience artifacts constructed in one agent framework are evaluated in another. Dashed lines indicate the target framework's Raw baseline.}
\label{fig:cross-agent}
\end{figure}

\begin{table}[t]
\centering
\small
\setlength{\tabcolsep}{4pt}
\renewcommand{\arraystretch}{1.12}
\begin{tabularx}{\columnwidth}{@{}p{0.32\columnwidth}L@{}}
\toprule
\textbf{Mechanism} & \textbf{Meaning} \\
\midrule
\texttt{procedural\_anchor} & The artifact gives a usable procedure, ordering, checklist, tool sequence, or verification plan. \\
\texttt{knowledge\_ injection} & The artifact supplies concrete domain knowledge that the agent otherwise lacked. \\
\texttt{failure\_warning} & The artifact warns about a pitfall that the agent avoids. \\
\texttt{none} & The artifact is not used in a meaningful way. \\
\texttt{counterproductive} & The artifact misleads the agent or makes the run worse. \\
\bottomrule
\end{tabularx}
\caption{\textbf{Mechanism labels used to characterize how injected prior experience affects execution.}}
\label{tab:taxonomy-mechanisms}
\end{table}

\section{Skill-Use Mechanisms: A Contrastive Taxonomy}



Aggregate success rates show whether skills help, but do not explain what changes when a skill is available. To make these changes observable, we build a contrastive trajectory-analysis pipeline over the controlled skill-vs-workflow experiments. We first normalize heterogeneous benchmark outputs into a shared manifest of 8,135 trial records, including task identity, execution arm, verifier outcome, injected artifact and trajectory transcript when available. Among these records, 7,837 contain agent transcripts. We then perform an open-coding pass over 240 sampled trajectories, retain 238 valid unique labels, and merge the resulting open-ended failure and success descriptions into a 12-mode canonical taxonomy.

We validate both LLM-assisted stages of this taxonomy construction with an independent human check. For each of the 238 valid raw labels, a human annotator inspected three supporting trajectories, giving 714 trajectory--label checks in total, and confirmed that the raw label was grounded in the recorded agent behavior. The annotator then independently mapped all 238 raw labels to the 12 canonical modes using only the taxonomy definitions. As summarized in Table~\ref{tab:taxonomy-human-validation}, the human and LLM aggregation assignments achieve 95.8\% exact agreement and Cohen's $\kappa=0.952$. This provides direct evidence that the taxonomy is stable beyond the judgments of a single LLM.

\begin{table}[t]
\centering
\small
\setlength{\tabcolsep}{4pt}
\renewcommand{\arraystretch}{1.10}
\begin{tabularx}{\columnwidth}{@{}>{\raggedright\arraybackslash}p{0.27\columnwidth}>{\raggedright\arraybackslash}p{0.35\columnwidth}>{\raggedright\arraybackslash}X@{}}
\toprule
\textbf{Validation stage} & \textbf{Evaluation units} & \textbf{Result} \\
\midrule
Trajectory grounding & 714 checks (238 labels $\times$ 3 trajectories) & All labels confirmed \\
Taxonomy aggregation & 238 valid unique labels & 95.8\% exact; Cohen's $\kappa=0.952$ \\
\bottomrule
\end{tabularx}
\caption{\textbf{Human validation of the taxonomy construction pipeline.} The first stage checks whether raw labels are supported by their source trajectories; the second independently maps those labels to the 12 canonical modes.}
\label{tab:taxonomy-human-validation}
\end{table}

The main unit of analysis is a paired triple. Each triple compares the same task and setting under three arms: raw execution, workflow-memory injection, and skill injection. We construct 528 such triples, covering SkillsBench (144), Terminal-Bench 2.0 (186), and Terminal-Bench-Pro (198). This gives 1,584 arm-level mode assignments. For each triple, an LLM judge assigns a taxonomy mode to each arm, records \textbf{pairwise changes between arms}, and identifies whether the injected artifact acts through procedural anchoring, knowledge injection, failure warning, no meaningful use, or counterproductive guidance. This paired design lets us ask not only whether an arm succeeds, but also which behavior changes when the same prior experience is represented as direct workflow memory or as a distilled skill.

At the coarse level, each trajectory is assigned a \textit{Skill-use Category} (SC), which serves as its top-level taxonomy label. The three SCs group the 12 fine-grained modes summarized in Table~\ref{tab:taxonomy-mode-percent}. SC1 captures successful \textit{\textbf{procedural anchoring}}, where the agent either succeeds autonomously or prior experience provides useful guidance. SC2 captures \textit{\textbf{execution-layer and verification failures}}, including environment setup, output formatting, service management, shell execution, algorithmic implementation, and runtime validation. SC3 captures \textit{\textbf{invocation, applicability, and boundary failures}}, where guidance is present but misused, over-applied, ignored, or constrained by external limits. At the group level, skill arms shift more trajectories into SC1 than workflow memory (326/528 skill-arm SC1 assignments vs. 294/528 for workflow memory), reduce SC2 execution-layer failures relative to raw and workflow memory (124/528 vs. 197/528 and 176/528), but also increase SC3 invocation or boundary failures (78/528 vs. 19/528 for raw). This organization directly decomposes the central question of the paper: when skills help, why they work, and where they fail.

\section{Findings}
\subsection{Skills work as procedural anchors, not merely external knowledge}

Skill-augmented runs achieve the highest oracle-status success rate, with 61.9\% success (oracle-status success rate) compared to 59.1\% for raw execution and 55.9\% for workflow memory. The strongest aggregate effect is therefore not skill over raw execution, but skill over direct workflow memory: skill improves over workflow memory by +6.06 percentage points, with a 95\% bootstrap confidence interval of [+0.76, +11.36]. This comparison is important because workflow memory and skill are constructed from the same source trajectories. The improvement therefore cannot be attributed merely to giving the agent more prior experience, but rather to how that experience is represented.

The taxonomy supports this interpretation. Under taxonomy-mode grouping, skill arms fall into the successful-procedure class in 61.7\% of cases (taxonomy-mode successful-procedure proportion), compared with 59.1\% for raw execution and 55.7\% for workflow memory. This is closely aligned with the oracle-status success rates, while allowing the judge to mark rare successful trajectories whose dominant behavior is still a failure-recovery pattern. More importantly, the mechanism labels show that skills mainly help through procedural anchoring: \texttt{procedural\_anchor} accounts for 65.7\% of skill mechanisms, whereas explicit \texttt{knowledge\_injection} accounts for only 4.5\%. Thus, skills usually do not work by supplying missing facts. They work by stabilizing action: which setup steps to run, which tool sequence to follow, what intermediate checks to perform, and which recurring pitfalls to avoid. A concrete matched trajectory example is provided in Appendix~\ref{app:qualitative-triple}.

The success modes make this distinction concrete. In the skill arm, \texttt{skill\_guided\_success} accounts for 61.6\% of cases. In the workflow arm, \texttt{workflow\_guided\_success} accounts for 54.5\%. This shows that workflow memory can also be useful: raw traces often contain reusable commands, parameter choices, and debugging evidence. However, workflow memory remains closer to the original trajectory and therefore preserves more incidental process, failed attempts, and task-specific details. Skills are more effective when they compress those traces into a cleaner operational procedure.

As an additional check that this effect is not simply caused by any compact procedural hint, we evaluate two lightweight baselines on the 26 selected Terminal-Bench-2 tasks: an instruction-derived short plan and a workflow-derived test-first template. These reach 47.7\% and 59.2\% success, respectively, below Workflow Memory (62.3\%) and substantially below Skill injection (79.2\%); full details are provided in Appendix~\ref{app:compact-baselines}. Appendix~\ref{app:token-cost} further shows an effectiveness--efficiency trade-off between Skill and Workflow Memory.

\subsection{Skills improve execution robustness but fall short on tasks requiring reformulation or verification}
The clearest practical advantage of skills appears in execution-layer and verification failures. SC2 modes account for 37.3\% of raw-arm labels and 33.3\% of workflow-arm labels, but only 23.5\% of skill-arm labels. This reduction supports the claim that skills are especially useful for operational fragility: they help agents avoid repeated setup mistakes, preserve output constraints, manage services more reliably, and use more robust command patterns.

The strongest example is \texttt{environment\_ infrastructure\_failure}, which drops from 5.3\% in raw execution to 1.7\% with workflow memory and 0.2\% with skills. This pattern suggests that environment and tooling problems are highly skillable. Once a reliable setup sequence, dependency workaround, or path convention has been discovered, it can be encoded as reusable procedural guidance. Similar reductions appear in \texttt{output\_format\_schema\_mismatch}, which decreases from 7.4\% in raw execution to 3.2\% with skills, and \texttt{background\_ service\_lifecycle\_failure}, which decreases from 2.7\% to 0.8\%. These failures are not usually caused by missing high-level reasoning; they arise when the agent fails to keep concrete execution constraints active during the run.

However, the taxonomy also shows the boundary of procedural guidance. \texttt{algorithmic\_ logic\_error} remains substantial across arms, at 8.3\% for raw execution, 11.0\% for workflow memory, and 7.4\% for skills. \texttt{static\_verification\_without\_runtime} is similarly persistent, at 12.5\% for raw execution, 12.5\% for workflow memory, and 11.7\% for skills. These modes show that skills do not automatically repair a wrong algorithm or force oracle-aligned validation. Skills improve execution robustness, but they do not eliminate failures that require deeper problem reformulation or stronger runtime verification.

\subsection{Skills introduce invocation and applicability failures}
The same abstraction that makes skills useful also creates a new failure surface. A skill is not self-executing: the agent must decide whether it applies, which parts to follow, how to adapt it, and when to abandon it. This is where many skill-specific failures arise. The mode \texttt{skill\_guidance\_misapplied\_or\_ignored} appears in 10.0\% of skill-arm cases, compared with only 0.8\% in raw execution and 0.4\% in workflow memory. These failures are not simply cases where the skill is absent or irrelevant. Often, the skill contains plausible guidance, but the agent applies it mechanically, misses a condition, or carries over assumptions that no longer hold.

Workflow memory fails differently. Its main penalty is not misapplication of a compact abstraction, but process overload. \texttt{timeout\_budget\_exhaustion} appears in 10.6\% of workflow-memory cases, compared with 1.7\% in raw execution and 4.4\% with skills. This suggests that direct traces can burden the agent with too much procedural residue: long explorations, failed attempts, and low-level debugging paths that distract from the decisive procedure. Skills reduce this overhead through distillation, but they do not eliminate the need for applicability judgment. A failed skill run may therefore reflect not bad skill content, but a failure to decide when and how the skill should govern the current execution.



Together, these patterns refine the interpretation of skill utility. A failed skill run is not always caused by bad skill content, just as a successful skill run is not always caused by the skill. Skill utility depends on a pipeline: prior experience must be distilled into the right level of abstraction, transferred or retrieved for a compatible context, invoked by the agent, and adapted during execution. The taxonomy therefore motivates our cross-framework and retrieval studies: we test whether distilled procedural guidance remains portable across agent frameworks and whether useful skills remain usable when retrieved from a larger or more confusable pool.

\subsection{Outcome annotations guide skill construction}
\begin{figure*}[t]
\centering
\includegraphics[width=0.85\textwidth]{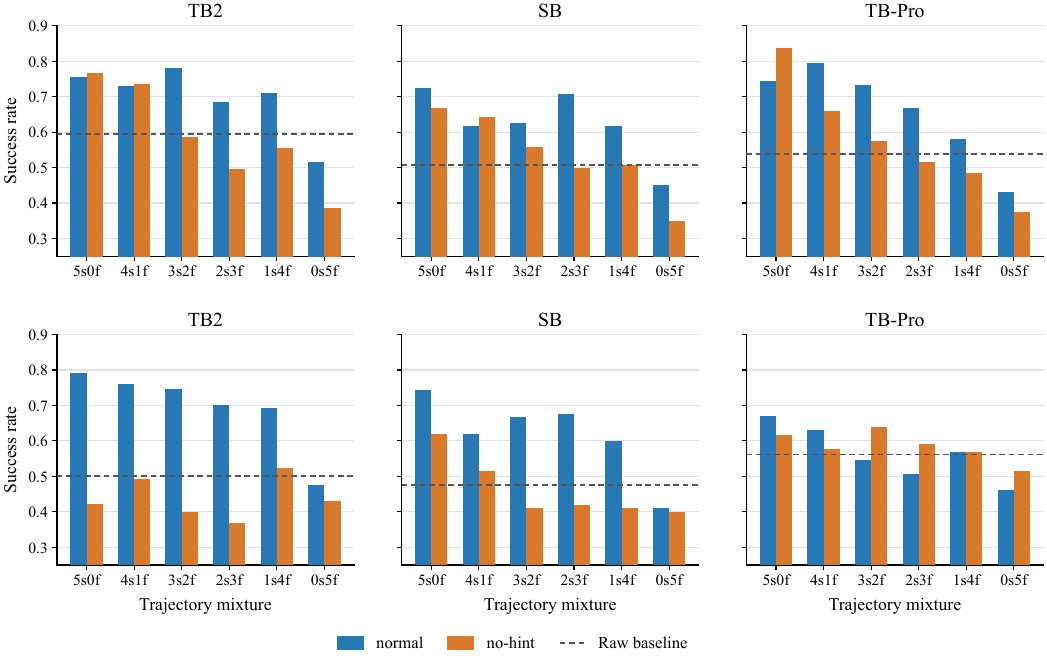}
\caption{\textbf{Effect of outcome labels during skill creation.} Panels compare skills created with outcome labels visible (\textit{normal}) or withheld (\textit{no-hint}) across trajectory mixtures and benchmarks. Dashed lines indicate the corresponding Raw baselines. Terminal-Bench-Pro entries use 130 trials per condition; missing or infrastructure-error trials count as failures.}
\label{fig:nohint-comparison}
\end{figure*}

Figure~\ref{fig:nohint-comparison} compares skills created from the same
trajectory pools with and without explicit success/failure labels for Codex and
Gemini CLI. The complete numerical results are reported in Appendix
Table~\ref{tab:nohint-skills-singlecol}.

For the completed Codex and Gemini rows, withholding outcome labels has little effect when the source pool contains only successful trajectories, but its impact grows sharply once failed trajectories are introduced. As failed trajectories enter the
pool, normal skill creation is generally stronger; for Gemini on
Terminal-Bench-2 at \texttt{3s2f}, it reaches 0.7462 versus 0.4000 without
outcome hints, and the same pattern holds across all completed Gemini
Terminal-Bench-2 and SkillsBench ratios.

\subsection{Retrieval exposes the gap between skill availability and skill use}
We evaluate retrieval on SkillsBench using the benchmark's ground-truth task--skill annotations. Each task is paired with a controlled candidate pool that contains the ground-truth skill and distractors. RQ4 comprises one downstream execution experiment and two independent offline retrieval diagnostics over matched pools. In the execution experiment, the complete pool is made available to the agent without preselecting skills; we parse the skills accessed during execution and measure their ground-truth overlap together with final task success. The offline diagnostics separately evaluate \textit{embedding-based retrieval}, which ranks skill descriptions against the task instruction with Qwen3-Embedding-0.6B, and \textit{agent selection}, which asks the agent to explicitly choose useful skills without executing the downstream task. The diagnostics do not provide inputs to the execution experiment.

Figure~\ref{fig:retrieval-combined}(b) first reports the independent downstream execution experiment. Gemini's parsed skill-use precision is low and decreases from 16.9\% to 0.7\%, while task success remains comparatively flat around 36--39\%. Codex starts with higher actual-use precision, 42.3\% at pool size 5, but also drops to 5.9\% at pool size 100; its task success instead increases from 35.4\% to 42.0\%. Averaged across the two reported pairings, downstream success changes only from 36.4\% to 39.3\% while actual-use precision falls from 29.6\% to 3.3\%. Moreover, Arm 3 recall remains 54.3--73.6\% at $k=100$, despite precision being only 0.7--8.1\%. This combination indicates that execution-time failure is not simply a failure to inspect any skill: agents often inspect or invoke multiple candidates, but do not reliably restrict use to the task's annotated ground-truth skill. Thus, exact ground-truth skill invocation is neither sufficient nor strictly necessary for success.

The two offline diagnostics show a different pattern. Figure~\ref{fig:retrieval-combined}(a) reports average top-1 embedding precision decreasing from 88.3\% at pool size 5 to 76.9\% at pool size 100, while explicit agent selection decreases from 70.0\% to 63.7\%. These values are independent measurements, not successive retrieval stages, and no selected skill is passed to the execution experiment. The offline results nevertheless locate the difficulty of identifying the relevant artifact before execution; the execution results above separately show what happens when the full pool is available during task solving.

The composition analysis identifies where offline identification becomes difficult. In Arm 1, top-1 precision on similar pools falls from 70.5\% at $k=5$ to 53.4\% at $k=100$, compared with 97.7\% to 84.1\% for random pools and 96.6\% to 93.2\% for dissimilar pools. The same asymmetry appears in explicit selection: at $k=5$, precision on similar pools is 54.3\% for Gemini and 51.9\% for Codex; at $k=100$, it is 55.4\% and 31.9\%, respectively. Thus, pool size contributes to the difficulty, but semantic confusability is the more important stressor for identifying the correct procedural artifact. The relatively high selection recall at $k=100$ (70.5--85.2\% across the reported conditions) further suggests that agents often include the ground-truth skill together with distractors rather than failing to consider it at all. A correct skill must ultimately be identified, noticed, adapted, and operationalized, while related non-ground-truth skills may still provide partial procedural support.

Table~\ref{tab:retrieval-stress-summary} reports the complete size trend behind the figure. It makes explicit that similar distractors are the dominant stressor for the offline identification diagnostics, whereas Arm 3 precision collapses across all three regimes as the pool grows. Downstream success is comparatively insensitive to this collapse, reinforcing that exact skill-use matching and task completion are independent measurements of different aspects of skill use rather than successive stages of one pipeline.

\begin{table}[t]
\centering
\scriptsize
\setlength{\tabcolsep}{3.0pt}
\renewcommand{\arraystretch}{1.02}
\begin{tabular}{@{}llccccc@{}}
\toprule
\multicolumn{2}{c}{\textbf{Pool / metric}} & \multicolumn{5}{c}{\textbf{Skill-pool size $k$}} \\
\cmidrule(lr){3-7}
\rowcolor{HeaderGray}
 & & \textbf{5} & \textbf{10} & \textbf{20} & \textbf{50} & \textbf{100} \\
\midrule
\multirow{4}{*}{\textit{Random}} & \cellcolor{PromptBlue}\textbf{Arm 1 P} & 97.7 & 95.5 & 95.5 & 92.0 & 84.1 \\
 & \cellcolor{ArmTwoBg}\textbf{Arm 2 P} & 78.1 & 77.9 & 82.1 & 76.5 & 69.8 \\
 & \cellcolor{ArmThreeBg}\textbf{Arm 3 P} & 25.9 & 23.2 & 19.5 & 8.6 & 4.4 \\
 & \cellcolor{TraceOutcomeBg}\textbf{Arm 3 Succ.} & 31.8 & 36.8 & 40.1 & 36.3 & 41.9 \\
\midrule
\multirow{4}{*}{\textit{Similar}} & \cellcolor{PromptBlue}\textbf{Arm 1 P} & 70.5 & 63.6 & 60.2 & 56.8 & 53.4 \\
 & \cellcolor{ArmTwoBg}\textbf{Arm 2 P} & 53.1 & 52.9 & 47.1 & 48.6 & 43.7 \\
 & \cellcolor{ArmThreeBg}\textbf{Arm 3 P} & 34.5 & 22.3 & 15.7 & 7.3 & 3.7 \\
 & \cellcolor{TraceOutcomeBg}\textbf{Arm 3 Succ.} & 41.7 & 39.6 & 39.2 & 39.5 & 39.6 \\
\midrule
\multirow{4}{*}{\textit{Dissimilar}} & \cellcolor{PromptBlue}\textbf{Arm 1 P} & 96.6 & 96.6 & 96.6 & 94.3 & 93.2 \\
 & \cellcolor{ArmTwoBg}\textbf{Arm 2 P} & 78.9 & 81.6 & 82.6 & 78.4 & 77.8 \\
 & \cellcolor{ArmThreeBg}\textbf{Arm 3 P} & 28.6 & 19.2 & 9.0 & 4.4 & 1.7 \\
 & \cellcolor{TraceOutcomeBg}\textbf{Arm 3 Succ.} & 35.7 & 36.9 & 33.7 & 38.8 & 36.4 \\
\bottomrule
\end{tabular}
\caption{\textbf{Effect of pool composition and size on SkillsBench.} Entries report percentages for the indicated retrieval arm, pool composition, and pool size. Arms 2 and 3 are arithmetic means over the reported agent--model pairings. Full recall and F1 values are given in Appendix Table~\ref{tab:retrieval-full-arm2-arm3}.}
\label{tab:retrieval-stress-summary}
\end{table}

\section{Conclusion}
This paper studies the behavior of skills through controlled experiments and contrastive trajectory analysis, moving beyond aggregate success rates to ask when skills help, why they work, and where they fail. Our results show that skills are most effective as procedural anchors and can also fail when they are retrieved incorrectly, invoked in the wrong context, followed too rigidly, or used on tasks that require deeper reformulation and runtime validation.

Overall, our findings suggest that skill use should be understood as a lifecycle problem rather than a single memory-injection mechanism. Building better self-evolving agents requires not only generating more skills, but also improving how agents represent, retrieve, and leverage procedural knowledge. We hope this analysis provides a foundation for more principled evaluation and design of future skill-based agent systems.

\section{Limitations}
Our evaluation focuses on terminal- and tool-using benchmarks that emphasize multi-step execution, debugging, and verification, and therefore does not cover the full range of agentic behavior, such as long-horizon web interaction or open-ended collaboration. It also evaluates a limited number of agent--model configurations, so the findings may not generalize to other scaffolds, model families, or model versions; we will extend the study to broader settings in future work. Finally, the mechanism taxonomy is derived from a stratified open-coding sample covering approximately 3\% of the normalized records rather than exhaustive labeling, so rare behavioral modes may be underrepresented.

\bibliography{custom}

\clearpage
\onecolumn
\appendix

\section{Implementation Details}
\label{app:experimental-details}

This part provides additional details for the four research questions described in the main text. Across the experiments, we use controlled comparisons designed to isolate representation, outcome annotation, cross-framework transfer, and retrieval difficulty. Unless otherwise stated, all conditions within an experiment share the same target tasks, benchmark interface, agent framework, model, execution harness, and trial budget.

\subsection{Model--Framework Pairings and Benchmarks}
\label{model_details}

We evaluate skill use in both realistic agent-framework deployments and controlled retrieval settings. For RQ1 and RQ2, the skill-vs-procedural-memory and no-hint protocols are implemented for two agent--model pairings: Codex + GPT-5.3-Codex and Gemini CLI + Gemini-3.1-Pro-Preview. RQ3 transfers workflow memories and skills constructed in the primary Codex setting to Gemini CLI with Gemini-3.1-Pro-Preview. RQ4 uses Qwen3-Embedding-0.6B for embedding retrieval and evaluates explicit agent selection and end-to-end execution with Gemini CLI + Gemini-3.1-Pro-Preview and Codex + GPT-5.4. GPT-5.3-Codex was no longer available under the same evaluation access when RQ4 was conducted, so GPT-5.4 was used for its Codex pairing. RQ4 is therefore interpreted only through within-pairing comparisons across its three independent experiments, rather than through direct numerical comparison with RQ1--RQ3.

Our evaluation suite combines Terminal-Bench~\cite{merrill2026terminalbenchbenchmarkingagentshard} and SkillsBench~\cite{li2026skillsbenchbenchmarkingagentskills}. RQ1--RQ3 use controlled subsets drawn from the public split of Terminal-Bench Pro, Terminal-Bench 2.0, and SkillsBench. Terminal-Bench Pro provides a 200-task public split from a 400-task benchmark spanning 8 domains; Terminal-Bench 2.0 contains 89 terminal-agent tasks; and SkillsBench contains 86 tasks across 11 domains designed for skill-based procedural reuse. RQ4 uses SkillsBench because it provides native task--skill annotations needed to evaluate retrieval precision and recall. These benchmarks are well suited to our setting because they require multi-step execution, tool use, debugging, service management, output validation, and runtime verification, making many failures procedural rather than purely factual.

Downstream execution experiments follow Harbor's standard evaluation workflow~\cite{Harbor_Framework_Team_Harbor_A_framework_2026}, with $n=5$ unique trials per task and a parallelism of $20$ unless otherwise noted. Retrieval-isolation arms use one query per task--pool setting because no benchmark execution is performed. For all skill-based execution conditions, we follow a standard agent-skill usage protocol in which skills are placed in the agent's execution environment as reusable procedural resources, rather than fully inlined into the initial context~\citep{anthropic_agent_skills}. For RQ4, semantic similarity scores used for retrieval and distractor construction are calculated with Qwen3-Embedding-0.6B~\citep{zhang2025qwen3embedding}, a 0.6B-parameter text embedding model from the Qwen3 Embedding series.

\subsection{Dive into Skill-Use Mechanisms: Trajectory Labeling and Comparative Analysis}
The analysis is designed to answer a mechanism-level question: \emph{when a prior-experience artifact is injected into an agent, what changes in the resulting trajectory and which failure modes are fixed or introduced?} Unlike aggregate success-rate evaluation, this pipeline treats each agent run as an executable trace. It aligns raw, workflow-memory and skill-injected executions for the same task, asks an LLM judge to compare the trajectories using a fixed taxonomy, and then aggregates the resulting paired labels into mode-level statistics. The complete prompts used for this part can be found in Appendix \ref{prompts}.

\paragraph{Trajectory Source and Experimental Context}
The trajectory corpus is derived from the controlled skill-vs-procedural-memory experiments. All trials were executed through the same benchmark harness, using a fixed agent-model configuration and a fixed task interface. The experiments cover three benchmark sources: Terminal-Bench 2.0, SkillsBench, and Terminal-Bench-Pro. As shown in Table \ref{tab:taxonomy-arms}, the agent is evaluated under three execution arms for each selected task: \textit{Raw}, \textit{Workflow memory} and \textit{Skill}. The workflow and skill arms are evaluated under six prior-experience compositions, ranging from all successful trajectories to all failed trajectories. These settings allow our analysis to observe how the quality of the underlying experience pool changes the downstream failure modes.

\begin{table}[t]
\centering
\small
\setlength{\tabcolsep}{4pt}
\renewcommand{\arraystretch}{1.12}
\begin{tabularx}{\columnwidth}{@{}p{0.23\columnwidth}LL@{}}
\toprule
\textbf{Arm} & \textbf{Injected prior experience} & \textbf{Purpose} \\
\midrule
Raw & No injected prior trajectory or skill & Baseline behavior of the agent on the task. \\
Workflow memory & Cleaned prior workflows are appended as procedural memory & Tests whether direct trajectory-like procedural memory improves execution. \\
Skill & The same prior workflows are distilled into a standardized reusable skill & Tests whether compact skill representation improves over direct workflow memory. \\
\bottomrule
\end{tabularx}
\caption{\textbf{Experimental arms in the contrastive trajectory analysis.}}
\label{tab:taxonomy-arms}
\end{table}

\begin{table}[t]
\centering
\small
\setlength{\tabcolsep}{4pt}
\renewcommand{\arraystretch}{1.12}
\begin{tabularx}{\columnwidth}{@{}p{0.30\columnwidth}L@{}}
\toprule
\textbf{Artifact} & \textbf{Role in the analysis} \\
\midrule
Trial result metadata & Stores task identity, reward, verifier result, exception type, timestamps, token usage, and execution phase durations. \\
Agent trajectory transcript & Stores the terminal/tool-use trajectory and the agent's reasoning-visible interaction record. \\
Task instruction & Defines the task objective and, for workflow-memory arms, may include injected workflow content. \\
Skill artifact & Stores the injected skill used in the skill arm. \\
Task-side files & Used only as contextual artifacts when present; the main taxonomy labels are based on execution trajectories and verifier outcomes. \\
\bottomrule
\end{tabularx}
\caption{\textbf{Input artifacts used by the taxonomy pipeline.}}
\label{tab:taxonomy-artifacts}
\end{table}

To make the analysis tractable, the raw execution bundle in Table \ref{tab:taxonomy-artifacts} is not fully mapped to the labeling prompt. Instead, the pre-processing step selectively extracts only the artifacts needed for trajectory analysis: verifier results, agent transcripts, task instructions, injected skills, and task-side configuration or solution metadata when available. This reduces the working corpus from a large raw execution archive to a compact analysis subset while preserving the evidence needed to explain success and failure. After normalization, the manifest contains 8,135 trial records. The distribution can be found in Table \ref{tab:taxonomy-manifest-coverage}. The manifest intentionally preserves records even when some auxiliary artifacts are missing. Later labeling stages filter to trials with sufficient evidence, especially an available trajectory transcript.

\begin{table}[t]
\centering
\small
\setlength{\tabcolsep}{5pt}
\renewcommand{\arraystretch}{1.08}
\begin{tabularx}{0.5\columnwidth}{@{}Lr@{}}
\toprule
\textbf{Dimension} & \textbf{Count} \\
\midrule
Terminal-Bench 2.0 trials & 3,254 \\
Terminal-Bench-Pro trials & 2,993 \\
SkillsBench trials & 1,888 \\
Raw-arm trials & 1,883 \\
Workflow-memory trials & 2,658 \\
Skill-arm trials & 3,594 \\
Successful trials & 4,541 \\
Failed trials & 3,594 \\
Records with available agent transcript & 7,837 \\
Records with available task instruction & 6,210 \\
Skill-arm records with linked skill artifact & 3,570 \\
\bottomrule
\end{tabularx}
\caption{\textbf{Coverage statistics after manifest construction and artifact linking.}}
\label{tab:taxonomy-manifest-coverage}
\end{table}

\paragraph{Manifest Construction and Trial Normalization}
The first stage converts heterogeneous benchmark outputs into a unified trial table. Each trial record stores the benchmark, task name, experimental setting, execution arm, reward, exception type, model identifier, timestamps, token metrics, and pointers to the relevant artifacts.

Success and failure are derived from the verifier reward rather than from free-form logs. A positive numeric verifier reward is treated as success; zero, missing, or non-positive reward is treated as failure. This design keeps the taxonomy aligned with the benchmark oracle rather than the agent's self-assessment.

The manifest builder also resolves the relationship between injected artifacts and trial outputs. For workflow-memory trials, the task instruction may contain the injected workflow memory. For skill trials, the corresponding skill artifact is linked by benchmark, setting, and task. This alignment is necessary because the later LLM judge must see not only what the agent did, but also what prior-experience artifact was available to it.

The implementation records token and duration metadata when present. This enables secondary analyses of context cost, output length, and execution time, although some skill-arm token fields are incomplete in the current exported result metadata. For this reason, the taxonomy analysis relies primarily on trajectory content, verifier outcomes, and paired mode changes.

\paragraph{Open Coding of Individual Trajectories}
Before applying a fixed taxonomy, the pipeline performs an open-coding stage over sampled individual trajectories. The goal is to induce a vocabulary from actual execution failures rather than impose a generic taxonomy from unrelated settings.

The sampler draws from cells defined by benchmark, setting, arm, and outcome. The default configuration targets 240 trials, with a minimum number of examples per cell and a fixed random seed. This stratification prevents the initial label pool from being dominated by the largest benchmark or by one execution arm.

Each sampled trajectory is passed to Claude Sonnet 4.6 through a headless command-line interface. Tool use is disabled and session persistence is disabled. These choices reduce contamination across labeling calls and force the model to judge only the evidence included in the prompt. 

The context budget is fixed before labeling. For individual-trial open coding, the task instruction is truncated to 3000-characters, the skill artifact to 3,000 characters, and the trajectory transcript to a head of 6,000 characters plus a tail of 12,000 characters. The larger tail budget reflects the empirical observation that final errors, verifier-facing decisions, and timeout behavior are usually concentrated near the end of the trajectory. Each LLM call has a 600-second timeout and is cached by trial id so interrupted or resumed runs do not relabel completed trials.

The output schema asks for a short explanation, an open-ended primary mode candidate, secondary factors, evidence spans, skill-effect judgment, and a coarse distinction among missing knowledge, misused knowledge, capability limit, and environmental failure. The open-ended primary\_mode\_candidate field is critical: it allows the initial vocabulary to emerge from the data.

This stage produced 240 raw-label records, of which 238 were retained as valid unique labels for taxonomy induction.

\paragraph{Canonical Taxonomy Induction}
The second stage converts open-ended labels into a canonical taxonomy. Because a single prompt over all labels would be long and brittle, the pipeline uses a two-round batched induction process.

First, labels are divided into batches of approximately 60 records. For each batch, the LLM proposes a small set of canonical modes and assigns each trajectory to exactly one mode. The prompt asks for specific procedural patterns rather than generic catch-all categories. It explicitly encourages coverage of environment failures, API or library misuse, debugging loops, verification mismatches, timeout, skill-specific behavior, workflow-specific behavior, and successful execution.

Second, batch-level modes are merged into a unified taxonomy. The merge prompt receives only the batch-level mode names, definitions, and counts, then produces a global set of modes and a mapping from every batch mode to one global mode. The local aggregation code then remaps individual assignments deterministically and validates that every trajectory id is assigned once.

The batch prompt asks for 8-14 modes per batch, and the merge prompt asks for 9-14 global modes. This range is a design choice: it prevents the taxonomy from collapsing into overly broad categories, such as the "wrong answer," while also avoiding a long tail of one-off labels that would not support statistical aggregation. The final merge produced 12 canonical modes.

This process yields a taxonomy with 12 modes. The modes are intentionally mixed success/failure modes: the goal is not only to classify why runs fail, but also to distinguish when success is attributable to skill guidance, workflow guidance, or autonomous agent behavior.

\paragraph{Paired Contrastive Trajectory Labeling}
The main analysis stage compares the trajectories matched for the same task. For each benchmark, setting, and task, the pipeline constructs a triple containing one raw trajectory, one workflow-memory trajectory, and one skill-injected trajectory whenever all three are available. Raw trials are matched by benchmark and task because raw execution does not have a success/failure mixture setting. Workflow and skill trials are matched by benchmark, task, and setting.

The current paired dataset contains 528 triples. As shown in Table \ref{tab:taxonomy-paired-sample}, every paired record includes all three arms. For each arm, the LLM judge receives the task instruction, the v1 taxonomy, the trial outcome, an excerpt of the agent trajectory, and the injected skill for the skill arm. The prompt asks the judge to output a structured comparison with three levels of information.

\begin{table}[t]
\centering
\small
\setlength{\tabcolsep}{5pt}
\renewcommand{\arraystretch}{1.08}
\begin{tabularx}{0.5\columnwidth}{@{}Lr@{}}
\toprule
\textbf{Split} & \textbf{Count} \\
\midrule
SkillsBench triples & 144 \\
Terminal-Bench 2.0 triples & 186 \\
Terminal-Bench-Pro triples & 198 \\
Triples per mixture setting & 88 \\
Total triples & 528 \\
\bottomrule
\end{tabularx}
\caption{\textbf{Paired triple sample used for contrastive taxonomy labeling.}}
\label{tab:taxonomy-paired-sample}
\end{table}

First, the judge assigns a v1 mode to each arm and provides an evidence quote. Second, it compares workflow with raw, skill with raw, and skill with workflow. Each comparison records a summary written in natural-language, a categorical net effect, the modes fixed by the treatment, and the modes introduced by the treatment. Third, it labels the mechanism by which the skill or workflow affected execution (Table \ref{tab:taxonomy-mechanisms}).

The paired-comparison prompt uses a shorter transcript budget than the individual open-coding prompt because it must include up to three trajectories in the same call. The task instruction and the injected skill are each truncated to 3,000 characters, while each trajectory contributes a 4,000-character head and an 8,000-character tail. This preserves the same head-to-tail evidence structure while fitting raw, workflow, and skill evidence into one comparison prompt.

Representative trial selection is deterministic by default: when multiple trials are available for the same task, setting, and arm, the first trial is selected under a stable trial-id ordering. A seeded random policy is also supported for sensitivity checks. The main run uses the deterministic policy to make the paired label set reproducible.

This paired design is the central methodological choice. Instead of asking which failure modes are frequent in isolation, it asks what changed when the representation of prior experience changed. This supports claims such as "\textit{skill fixed environment setup failures that raw execution encountered}" or "\textit{skill introduced a misapplication failure that was absent in raw execution}".

\subsection{Paired-Trajectory Example}
\label{app:qualitative-triple}

The following excerpts show Raw, Workflow Memory, and Skill executions for the
same SkillsBench task, \texttt{react-performance-debugging}, under the
\texttt{1s4f} setting. Unrelated file inspection and repeated build output are
omitted.

\begingroup
\traceheading{TraceRawFrame}{Raw execution \quad (reward 0)}
\begin{lstlisting}[style=trace-raw]
[Agent]
I hit an unexpected regression: /api/products is
returning 500 in the production run. I am checking
server logs now, then I will patch the route so it
stays fast without breaking correctness.

[Runtime checks]
api/products run1 total=0.018289 status=500
api/products run2 total=0.014010 status=500
api/checkout run1 total=0.005307 status=200
api/checkout run2 total=0.003592 status=200

[Agent]
I found a runtime issue in production mode: client
hydration is not completing. I am rebuilding and
restarting the production server from the latest source.

$ npm run start
Error: listen EADDRINUSE: address already in use :::3000

[Verifier]
test_checkout_fast                 PASSED
test_external_api_actually_called  FAILED
AssertionError: Checkout API too fast (7 ms) -
external API may be bypassed
Result: 10 passed, 1 failed
\end{lstlisting}

\traceheading{TraceWorkflowFrame}{Workflow Memory \quad (reward 0)}
\begin{lstlisting}[style=trace-workflow]
[Agent]
I found the main hotspots: server-side waterfalls on the
homepage and product API route, blocking analytics in the
API response path, and a very heavy compare page bundle.
I am now implementing targeted fixes.

[Final checkout patch]
export async function POST() {
  const [user, config] = await Promise.all([
    fetchUserFromService(),
    fetchConfigFromService(),
  ]);

  const profile =
    await fetchProfileFromService(user.id);

  return NextResponse.json({
    success: true,
    user: { id: user.id, name: user.name },
    profile,
    config: { currency: config.currency },
  });
}

[Verifier]
test_external_api_actually_called  PASSED
test_checkout_fast                 FAILED
AssertionError: Checkout took 915 ms
(should be <800 ms)
Result: 10 passed, 1 failed
\end{lstlisting}

\traceheading{TraceSkillFrame}{Skill \quad (reward 1)}
\begin{lstlisting}[style=trace-skill]
[Injected SKILL.md]
Eliminate server/API waterfalls:
- Convert independent awaits to Promise.all.
- Start promises early, await late.
- For partially dependent flows, fetch independent data
  in parallel, then trigger the dependent fetch as soon
  as its prerequisite resolves.

[Final checkout patch]
export async function POST() {
  const userPromise = fetchUserFromService();
  const configPromise = fetchConfigFromService();
  const user = await userPromise;
  const profilePromise =
    fetchProfileFromService(user.id);
  const [config, profile] = await Promise.all([
    configPromise,
    profilePromise,
  ]);

  return NextResponse.json({
    success: true,
    user: { id: user.id, name: user.name },
    profile,
    config: { currency: config.currency },
  });
}

[Agent runtime measurement]
POST /api/checkout:
0.722233, 0.709605, 0.713089, 0.710551, 0.713057 s
warm average = 0.711576 s

[Verifier]
test_checkout_fast                 PASSED
test_external_api_actually_called  PASSED
test_cart_add_item                 PASSED
test_compare_page_works            PASSED
Result: 11 passed in 11.74 s
\end{lstlisting}
\endgroup

\traceoutcome{Raw bypasses the external-service check, Workflow Memory leaves the dependent profile request serialized and exceeds the latency threshold, and Skill starts that request as soon as its prerequisite resolves and passes all 11 tests.}

\paragraph{Deterministic Aggregation and Statistical Reporting}
The final stage aggregates the paired labels without additional LLM calls. It computes per-arm success rates, paired success-rate deltas, mode frequencies, mode-level fixed/introduced counts, mechanism distributions, setting-level trends, benchmark-level trends, and token/duration summaries.

Paired success-rate deltas are computed per triple and summarized with a 1,000-iteration bootstrap confidence interval. As shown in Table \ref{tab:taxonomy-paired-deltas}, the most robust aggregate difference in this taxonomy sample is not skill versus raw, but skill versus direct workflow memory. This is consistent with the paper's broader claim: \emph{skills are useful because they distill prior trajectories into a more compact and actionable form, while workflow memory can preserve too much noisy process}. 

\begin{table}[t]
\centering
\small
\setlength{\tabcolsep}{5pt}
\renewcommand{\arraystretch}{1.08}
\begin{tabularx}{0.8\columnwidth}{@{}Lrr@{}}
\toprule
\textbf{Arm} & \textbf{Success / total} & \textbf{Success rate} \\
\midrule
Raw & 312 / 528 & 59.1\% \\
Workflow memory & 295 / 528 & 55.9\% \\
Skill & 327 / 528 & 61.9\% \\
\bottomrule
\end{tabularx}
\caption{\textbf{Oracle-status success rates across the three execution arms.}}
\label{tab:taxonomy-success-rates}
\end{table}

\begin{table}[t]
\centering
\small
\setlength{\tabcolsep}{4pt}
\renewcommand{\arraystretch}{1.08}
\begin{tabularx}{0.8\columnwidth}{@{}Lrr@{}}
\toprule
\textbf{Comparison} & \textbf{Mean paired delta} & \textbf{95\% bootstrap CI} \\
\midrule
WM vs Raw & $-0.0322$ & $[-0.0814, +0.0208]$ \\
Skill vs Raw & $+0.0284$ & $[-0.0227, +0.0795]$ \\
Skill vs WM & $+0.0606$ & $[+0.0076, +0.1136]$ \\
\bottomrule
\end{tabularx}
\caption{\textbf{Paired success-rate deltas between execution arms.}}
\label{tab:taxonomy-paired-deltas}
\end{table}

\paragraph{Reproducibility and Reliability Controls}
Several design choices make the pipeline reproducible and auditable.
The manifest uses deterministic parsing rules for benchmark, setting, arm, task, and reward. The sampling stage uses a fixed seed and stratified cells. The taxonomy induction stage caches intermediate LLM outputs and validates that every input id is assigned exactly once. The paired-comparison stage caches each task-setting comparison, supports fixed representative-trial selection, and includes a kill switch to avoid producing long runs of invalid labels under rate limits. The final report is deterministic and does not use LLM calls.

The LLM judge is constrained by strict JSON schemas and evidence quotes. Evidence quotes are important because they make a label inspectable: each mode assignment should be traceable to the trajectory, instruction, result metadata, or skill artifact. The paired prompt also exposes the same task under multiple arms, reducing the risk that the judge attributes a failure to skill use when the same failure also appears in raw execution.

There are also limitations. The v1 taxonomy is induced from 238 valid unique labels, then applied to 528 paired triples. Although the trajectory grounding and taxonomy aggregation are independently human-validated (Table~\ref{tab:taxonomy-human-validation}), the full paired-triple dataset remains LLM-assisted rather than exhaustively human-coded. The representative trial policy selects one trajectory per arm for each task-setting triple, so it does not average all repeated trials. Token metrics are incomplete for some skill-arm runs; we therefore report token cost only on a matched same-task intersection with complete metadata (Appendix~\ref{app:token-cost}), while treating mechanism and mode analyses as the primary evidence for behavioral interpretation.

\subsection{RQ1: Representation of Prior Experience}

RQ1 asks whether representing prior experience as a standardized skill differs from injecting the same experience as direct procedural memory. We compare three conditions: \emph{Raw}, \emph{Workflow Memory}, and \emph{Skill}. In Raw, the agent receives no prior experience. In Workflow Memory, the agent receives cleaned procedural memories derived from prior executions. In Skill, the same workflows are distilled into a standardized reusable skill. The central control is that workflow memories and skills are constructed from the same underlying trajectory pool; the manipulated variable is only how that experience is represented and made available to the agent.

We first collect raw terminal and tool-use trajectories on Terminal-Bench 2.0, SkillsBench, and Terminal-Bench Pro under a fixed execution protocol. Within each agent--model pairing, we keep the benchmark interface, task definition, model, agent scaffold, and trial budget constant. The same protocol is instantiated for Codex + GPT-5.3-Codex and Gemini CLI + Gemini-3.1-Pro-Preview. We retain tasks whose raw runs contain both successful and failed trajectories and continue collecting executions until each selected task has a balanced trajectory pool with sufficient successful and failed runs for controlled recomposition. This shared pool serves as the common source for all subsequent Workflow Memory and Skill conditions.

Workflow memories are constructed by cleaning and structuring the trajectories while preserving their procedural flow. Skills are constructed by distilling the same workflows into a standardized \texttt{SKILL.md} representation. Under a fixed experience budget, we systematically vary the composition of successful and failed trajectories and rerun the selected tasks under Raw, Workflow Memory, and Skill conditions. We primarily report task success rate and token/context cost. This setup isolates whether representing the same prior experience as direct procedural memory or as a distilled skill changes downstream execution.

\subsection{RQ2: Outcome Annotation and No-Hint Ablation}

RQ2 asks whether the benefits of skills come from the underlying experience itself or from explicitly exposing success/failure outcomes. Starting from the same selected tasks and balanced trajectory pools used in RQ1, we construct \emph{standard} and \emph{no-hint} Skill variants. In the standard setting, success/failure identities remain visible to the skill creator. In the no-hint setting, explicit outcome annotations are removed while the underlying trajectories, task pool, experience budget, and execution protocol remain unchanged.

For Skill, the no-hint variant withholds outcome annotations from the skill-construction stage while keeping the same workflow content and standardized skill format. We then rerun the same tasks under matched conditions and compare success rate and token/context cost across standard and no-hint settings. This design separates the effect of experience content from the effect of explicit annotation of results during skill construction.

\subsection{RQ3: Cross-Framework Transfer}

RQ3 evaluates whether reusable procedural knowledge is tied to the framework that produced it. We construct workflow memories and skills from trajectories collected with Codex and GPT-5.3-Codex, then evaluate those fixed artifacts with Gemini CLI and Gemini-3.1-Pro-Preview. The target tasks and source experience are held fixed, while the agent framework changes in prompting style, tool interface, and execution loop.

We compare transferred Workflow Memory and Skill against the Gemini raw baseline across the same trajectory-mixture settings. Because both artifacts originate from the same Codex trajectories, differences between them reflect how directly preserved workflow traces and distilled skills survive the framework shift. This design operationalizes portability as downstream task success under a new agent framework rather than textual similarity between artifacts.

\subsection{RQ4: Skill Retrieval and Downstream Execution}
\label{rq4}
To characterize skill identification and execution on SkillsBench, RQ4 comprises three independent experiments over matched candidate pools: two offline diagnostics and one downstream execution experiment. SkillsBench provides native ground-truth task--skill annotations, which lets us compute precision, recall, and F1 without using downstream task success as a circular proxy for relevance.

For each task $t$, we define the ground-truth skill set $G_t$ as the set of canonical skill identifiers attached to $t$ by the benchmark's native task--skill annotations. We do not infer $G_t$ from retrieval outputs, agent success, or the skill descriptions generated in our experiments. Candidate-pool duplicates are removed by canonical skill identifier. For a query or execution trial $i$, let $\widehat{G}_i$ be the set of distinct skills selected by the agent or extracted by the execution-time parser. A skill counts as correct only when its canonical identifier occurs in both sets; a useful distractor that is not in $G_t$ remains a false positive. Repeated mentions or invocations of the same skill count once.

In Arm 1, \emph{embedding-based retrieval}, we encode the task instruction and each skill description with Qwen3-Embedding-0.6B and rank skills by cosine similarity. The strict setting returns the single nearest skill and reports top-1 precision. We also retain top-$k$ retrieval statistics in the released artifacts, but the main figure reports the single-skill selection setting because it matches the question of whether the retriever can identify the best skill directly.

In Arm 2, \emph{explicit agent selection}, each task is presented with an \texttt{available\_skills} candidate pool and the agent must explicitly choose the skills it would use. The downstream task is not executed. This arm evaluates whether an agent can use task context and skill descriptions to choose helpful skills, without conflating selection with later execution failures. We run this selection protocol with Gemini CLI + Gemini-3.1-Pro-Preview and Codex + GPT-5.4.

In Arm 3, \emph{real execution}, the complete candidate pool is placed in the agent's execution environment and the agent runs the benchmark task without receiving a preselected skill from either offline diagnostic. After the run, we parse the trajectory to identify which skills were actually inspected or invoked. We then compute precision, recall, and F1 over parsed skill use, together with the benchmark success rate. This arm uses the same two agent--model pairings as Arm 2 and measures whether skills available in the environment are actually operationalized during execution.

For each valid query or trial record, precision and recall are computed from the predicted and gold sets as
\begin{equation}
P_i = \frac{|\widehat{G}_i \cap G_t|}{|\widehat{G}_i|}, \qquad
R_i = \frac{|\widehat{G}_i \cap G_t|}{|G_t|}.
\end{equation}
For a condition with $N$ valid records, we aggregate as
\begin{equation}
P = \frac{1}{N}\sum_{i=1}^{N} P_i, \qquad
R = \frac{1}{N}\sum_{i=1}^{N} R_i, \qquad
F_1 = \frac{2PR}{P+R}.
\end{equation}
Because every selected task has at least one annotated gold skill, $|G_t|>0$; an empty predicted set receives $P_i=R_i=0$, and $F_1$ is set to zero when $P+R=0$. For each pool-size and distractor condition, $P$ and $R$ are arithmetic means of the per-record values: Arm 1 and Arm 2 contribute one query record per task--pool setting, whereas Arm 3 contributes one record per task and trial. The reported $F_1$ is recomputed from the aggregated $P$ and $R$, rather than averaged from per-record F1 values, and task success is the mean of the binary verifier outcomes. When results are averaged across distractor regimes, the averages are computed from the unrounded condition-level values and rounded to one decimal place only for presentation.

All three experiments share the same candidate-pool construction. Each pool contains the task's ground-truth skill set and distractors. Pool size varies over 5, 10, 20, 50, and 100. Distractors are sampled under three regimes: random distractors from unrelated skills, similar distractors selected as embedding-space near-neighbors, and dissimilar distractors selected from far-away skills. Thus, the benchmark, task set, ground truth, seed, and pool-size schedule remain fixed; only the evaluation procedure and distractor composition change.

This design measures three complementary aspects of skill use: offline semantic identification, deliberate selection, and execution-time access under a full candidate pool. The experiments are not sequential: correct offline selection is not supplied to the execution run, and execution-time access is not treated as a prerequisite for the offline diagnostics. This distinction is important because correct selection is not guaranteed to produce successful execution, and incorrect exact-ground-truth selection is not always fatal: related non-ground-truth skills can still provide useful procedural guidance.

\begin{table}[t]
\centering
\scriptsize
\setlength{\tabcolsep}{3pt}
\renewcommand{\arraystretch}{1.08}
\begin{tabularx}{\columnwidth}{@{}p{0.10\columnwidth}Xccc@{}}
\toprule
\textbf{SC} & \textbf{Mode} & \textbf{Raw} & \textbf{WF} & \textbf{Skill} \\
\midrule
SC1 & \texttt{skill\_guided\_success} & 10.4\% & 0.4\% & 61.6\% \\
SC1 & \texttt{workflow\_guided\_success} & 0.0\% & 54.5\% & 0.0\% \\
SC1 & \texttt{autonomous\_clean\_success} & 48.7\% & 0.8\% & 0.2\% \\
\midrule
SC2 & \texttt{environment\_infrastructure\_failure} & 5.3\% & 1.7\% & 0.2\% \\
SC2 & \texttt{output\_format\_schema\_mismatch} & 7.4\% & 3.8\% & 3.2\% \\
SC2 & \texttt{background\_service\_lifecycle\_failure} & 2.7\% & 2.5\% & 0.8\% \\
SC2 & \texttt{shell\_code\_corruption} & 1.1\% & 1.9\% & 0.2\% \\
SC2 & \texttt{algorithmic\_logic\_error} & 8.3\% & 11.0\% & 7.4\% \\
SC2 & \texttt{static\_verification\_without\_runtime} & 12.5\% & 12.5\% & 11.7\% \\
\midrule
SC3 & \texttt{timeout\_budget\_exhaustion} & 1.7\% & 10.6\% & 4.4\% \\
SC3 & \texttt{skill\_guidance\_misapplied\_or\_ignored} & 0.8\% & 0.4\% & 10.0\% \\
SC3 & \texttt{capability\_or\_safety\_limit} & 1.1\% & 0.0\% & 0.4\% \\
\bottomrule
\end{tabularx}
\caption{\textbf{Contrastive skill-use taxonomy over 528 paired triples.} \textit{SC} abbreviates \textit{Skill-use Category}, the top-level taxonomy label assigned to a trajectory; each SC groups the fine-grained modes listed in the table. Percentages are computed within each arm over the same paired-triple sample. SC1 denotes successful procedural anchoring, SC2 execution-layer and verification failures, and SC3 invocation, applicability, and boundary failures.}
\label{tab:taxonomy-mode-percent}
\end{table}




\subsection{Lightweight Compact Procedural Baselines}
\label{app:compact-baselines}

To test whether the observed skill advantage can be explained by compact procedural text alone, we add two lightweight baselines on the same 26 selected Terminal-Bench-2 tasks used in the Gemini CLI + Gemini-3.1-Pro-Preview comparison. Each condition is evaluated with five trials per task, giving 130 trials. The \emph{short-plan} baseline provides a concise instruction-derived plan with three to five high-level steps. The \emph{test-first} baseline provides a workflow-derived validation template that emphasizes success conditions, intermediate checks, and final verification. Both baselines are injected as plain procedural text rather than as reusable \texttt{SKILL.md} artifacts.

\begin{table}[t]
\centering
\small
\setlength{\tabcolsep}{4pt}
\renewcommand{\arraystretch}{1.08}
\begin{tabularx}{\columnwidth}{@{}Lrrr@{}}
\toprule
\textbf{Condition} & \textbf{Source} & \textbf{Success / total} & \textbf{Success rate} \\
\midrule
Raw & None & 65 / 130 & 50.0\% \\
Short plan & Task instruction & 62 / 130 & 47.7\% \\
Test-first template & Workflow & 77 / 130 & 59.2\% \\
Workflow Memory & Workflow & 81 / 130 & 62.3\% \\
Skill & Workflow & \textbf{103 / 130} & \textbf{79.2}\% \\
\bottomrule
\end{tabularx}
\caption{\textbf{Lightweight compact procedural baselines on selected Terminal-Bench-2 tasks.} Entries report downstream success for Raw, short-plan, test-first, Workflow Memory, and Skill conditions over 26 tasks with five trials per task.}
\vspace{-5mm}
\label{tab:compact-procedural-baselines}
\end{table}

\begin{table*}[h]
\centering
\scriptsize
\setlength{\tabcolsep}{4.5pt}
\renewcommand{\arraystretch}{1.12}
\begin{tabularx}{\textwidth}{@{}lrrrr>{\raggedright\arraybackslash}p{0.14\textwidth}X@{}}
\toprule
\rowcolor{HeaderGray}
\multicolumn{7}{@{}l}{\textbf{Absolute metrics on the matched 83-task intersection}} \\
\textbf{Representation} &
\textbf{Success} &
\textbf{Input} &
\textbf{Output} &
\textbf{Total} &
\textbf{$\Delta$ succ. vs Raw} &
\textbf{Cost profile} \\
\midrule
Raw trajectories & 64.1\% & 541.5K & 14.2K & 555.7K & -- & Full prior traces provide broad evidence but carry the largest context load. \\
Workflow Memory & 64.8\% & 417.9K & 8.3K & 426.2K & +0.7 pp & Most token-efficient representation after cleaning trajectory noise. \\
Skill & 69.6\% & 511.7K & 9.8K & 521.5K & +5.5 pp & Highest success rate, with lower token use than Raw but higher token use than Workflow Memory. \\
\midrule
\rowcolor{HeaderGray}
\multicolumn{7}{@{}l}{\textbf{Pairwise trade-offs}} \\
\textbf{Comparison} &
\textbf{$\Delta$ success} &
\textbf{$\Delta$ input} &
\textbf{$\Delta$ output} &
\textbf{$\Delta$ total} &
\textbf{Direction} &
\textbf{Interpretation} \\
\midrule
Workflow Memory vs Raw & +0.7 pp & -123.6K & -5.9K & -129.5K & cheaper & Workflow Memory substantially reduces token cost with nearly unchanged success. \\
Skill vs Raw & +5.5 pp & -29.8K & -4.4K & -34.2K & better and cheaper & Skill improves success while still reducing token use relative to Raw trajectories. \\
Skill vs Workflow Memory & +4.8 pp & +93.8K & +1.5K & +95.3K & better but costlier & Skill trades additional context for stronger execution performance. \\
\bottomrule
\end{tabularx}
\caption{\textbf{Matched success and token-cost comparison.} Entries are computed on the 83-task intersection with equal task weighting. Token counts are per-task averages reported in thousands (K); ``pp'' denotes percentage points.}
\label{tab:matched-token-cost}
\end{table*}

\subsection{Matched Token-Cost Analysis}
\label{app:token-cost}

We report token usage on a matched intersection of 83 tasks for which Raw, Workflow Memory, and Skill runs all contain usable token metadata. To avoid over-weighting tasks with more completed trials, we first average success and token usage within each task and representation, then average across tasks. This same-task comparison keeps the task mix fixed when comparing representation choices.

The result reveals an effectiveness--efficiency trade-off. Workflow Memory is the most token-efficient representation, reducing both input and output tokens relative to Raw trajectories. Skill is not uniformly cheaper than Workflow Memory, but it achieves the highest success rate: it improves over Raw while also reducing token usage, and it trades additional context relative to Workflow Memory for stronger execution performance. We therefore interpret Skill as the more effective representation and Workflow Memory as the more token-efficient one.





\section{Prompts}
\label{prompts}

\subsection{skill-creator.md (Experiment 1)}

\begin{promptlisting}{Skill Creator Prompt}
You are a skill generator. Given one or more execution traces from an agent completing a task, you will produce a single reusable skill file that captures the repeatable process.

Analyze the traces to identify:
- What repeatable process was performed
- The distinct steps (in order)
- What tools, commands, and libraries were used
- Common patterns across traces (if multiple traces provided)
- Failure modes that appeared (if any failed traces are included)

Then write a skill in the following markdown format:

---
name: {{skill-name}}
description: {{one-line description}}
---

# {{Skill Name}}

## Use This Skill When
- {{condition 1}}
- {{condition 2}}

## Preconditions
- {{what must be true before starting}}

## Steps
1. {{step 1}}
2. {{step 2}}
...

## Common Failure Modes To Avoid
- {{failure mode 1: signal and mitigation}}
- {{failure mode 2: signal and mitigation}}

## If A Failure Happens
1. Stop and inspect the latest output.
2. Map the error to the failure modes above and apply the fix.
3. Re-run verification before finishing.

## Verify
- {{how to confirm the skill completed successfully}}

Rules:
- Produce exactly ONE skill, not multiple.
- The skill should be general enough to apply to similar tasks, not just the exact task in the traces.
- Steps should be concrete and actionable, not vague.
- If multiple traces show different approaches, pick the most reliable one.
- If failed traces are included, extract their failure patterns into the "Common Failure Modes" section.
- Do not include task-specific file paths or data -- use placeholders.
\end{promptlisting}

\subsection{skill-creator-no-hint.md (Experiment 2)}

\begin{promptlistingpurple}{No-Hint Skill Creator Prompt}
    You are a skill generator. Given one or more execution traces from an agent completing a task, you will produce a single reusable skill file that captures the repeatable process.

Analyze the traces to identify:
- What repeatable process was performed
- The distinct steps (in order)
- What tools, commands, and libraries were used
- Common patterns across traces (if multiple traces provided)
- Failure signals and mitigations inferred from observable evidence in traces

Then write a skill in the following markdown format:

---
name: {{skill-name}}
description: {{one-line description}}
---

# {{Skill Name}}

## Use This Skill When
- {{condition 1}}
- {{condition 2}}

## Preconditions
- {{what must be true before starting}}

## Steps
1. {{step 1}}
2. {{step 2}}
...

## Common Failure Modes To Avoid
- {{failure mode 1: signal and mitigation}}
- {{failure mode 2: signal and mitigation}}

## If A Failure Happens
1. Stop and inspect the latest output.
2. Map the error to the failure modes above and apply the fix.
3. Re-run verification before finishing.

## Verify
- {{how to confirm the skill completed successfully}}

Rules:
- Produce exactly ONE skill, not multiple.
- The skill should be general enough to apply to similar tasks, not just the exact task in the traces.
- Steps should be concrete and actionable, not vague.
- If multiple traces show different approaches, pick the most reliable one.
- Infer likely failure patterns only from observable evidence in the traces (e.g., command outputs, error text, exit status, retries).
- Do not assume whether any trace is successful or failed unless the evidence supports it.
- Do not include task-specific file paths or data -- use placeholders.

\end{promptlistingpurple}





\subsection{Prompts for Trajectory Labeling and Comparative Analysis}

\begin{promptlisting}{Trajectory Labeling Prompt}
You are labeling agent trajectories from a controlled study comparing three
conditions on a software-engineering benchmark:
  - raw:       no procedural memory, no skill injected
  - workflow:  past workflow memories appended to instruction
  - skill:     a curated SKILL.md injected into the agent's environment

The trial outcome is given (success/failure with a reward). Your job is to
explain WHY the trial ended that way, using the trajectory and (when relevant)
the candidate skill or injected workflow. Be concrete; cite tool calls or
output snippets.

Output STRICT JSON only (no preamble, no markdown fences) with this schema:
{
  "freeform_reasoning_short": "1-2 sentences summarizing what happened",
  "primary_mode_candidate": "short label, free text, e.g. 'missing python dependency' or 'agent looped on same failing test'",
  "secondary_factors": ["short label", "..."],
  "evidence_spans": [
    {"source": "codex.txt" | "result.json" | "instruction.md" | "SKILL.md",
     "quote": "verbatim snippet, <=160 chars"}
  ],
  "skill_effect_judgment": "helps" | "neutral" | "hurts" | "not_applicable",
  "skill_effect_reason": "one short sentence; for arm != skill use not_applicable + ''",
  "capability_vs_knowledge": "knowledge_missing" | "knowledge_present_but_misused" | "capability_limit" | "environmental"
}

Rules:
  - For arm=raw and arm=workflow trials, skill_effect_judgment MUST be "not_applicable".
  - "primary_mode_candidate" should be a noun phrase, not a sentence.
  - Always provide at least one evidence_span quoting the trajectory or result.
  - If trial succeeded, primary_mode_candidate should describe the success path
    (e.g. "clean python implementation passed all tests") and capability_vs_knowledge
    can be "knowledge_present_but_misused" only if the agent recovered from misuse.
\end{promptlisting}

\begin{promptlistingpurple}{Batch Mode Aggregation Prompt}
You are mining canonical failure/success modes from agent-trajectory labels.

Each input record (one trajectory) has:
  id, benchmark, setting, arm, status,
  primary_mode_candidate (free text), secondary_factors,
  skill_effect_judgment, capability_vs_knowledge.

Task: produce a small set of canonical modes (snake_case names) covering THIS BATCH,
and assign every input id to exactly one mode.

Modes should be specific procedural patterns, not generic catch-alls. Cover:
  - environment/dependency failures
  - API or library misuse
  - debugging-loop / no-progress
  - verification/output-format mismatches
  - timeout / long-horizon failures
  - skill-specific patterns (skill misguidance, skill ignored, skill helped)
  - workflow-specific patterns
  - successful-execution patterns (clean / recovered / skill-guided)

Output STRICT JSON only (no markdown fences):
{
  "modes": [
    {"name": "missing_python_dependency",
     "definition": "1-3 sentences",
     "n_assigned": 17}
  ],
  "assignments": [
    {"id": "<trial_id>", "mode": "missing_python_dependency", "reason": "one short sentence"}
  ]
}

Rules:
- Every input id MUST appear once in assignments.
- Mode names in assignments MUST be defined in modes.
- Aim for 8-14 modes per batch. Prefer fewer, clearer modes over many narrow ones.

Input records:
{INPUT_RECORDS}
\end{promptlistingpurple}

\begin{promptlisting}{Taxonomy Merge Prompt}
You are consolidating mode taxonomies from multiple batches into a single
canonical taxonomy.

Each batch proposed a list of modes (name + definition + n_assigned). Many
batch modes will overlap. Your job: produce a unified set of 9-14 canonical
modes, and a mapping from each batch-level mode name to its unified target.

Output STRICT JSON only (no markdown fences):
{
  "modes": [
    {"name": "unified_snake_case",
     "definition": "1-3 sentences",
     "merged_from_batch_modes": ["batch1_mode_name", "batch2_mode_name"]}
  ],
  "batch_mode_map": {
    "batch1_mode_name": "unified_snake_case",
    "batch2_mode_name": "unified_snake_case"
  }
}

Rules:
- Every batch mode name that appears in input MUST appear as a key in batch_mode_map.
- Every value in batch_mode_map MUST be a name in modes[].
- Aim for 9-14 unified modes total.
- "merged_from_batch_modes" lists the source-batch names that were folded into each unified mode.

Batch mode lists:
{BATCH_MODE_LISTS}
\end{promptlisting}

\begin{promptlistingpurple}{Paired Comparison Prompt}
You compare 3 agent trajectories on the same task across 3 conditions
(raw / workflow-injected / skill-injected) and produce a structured paired analysis.

---TASK INSTRUCTION (may contain injected workflow text if from workflow arm)---
{TASK_INSTRUCTION}

---SHARED MODE TAXONOMY (use these names in `mode` fields)---
{SHARED_MODE_TAXONOMY}

[ARM=raw] status={STATUS} reward={REWARD} exception={EXCEPTION_TYPE}
---codex.txt (head + tail)---
{RAW_CODEX_EXCERPT}

[ARM=workflow] status={STATUS} reward={REWARD} exception={EXCEPTION_TYPE}
---codex.txt (head + tail)---
{WORKFLOW_CODEX_EXCERPT}

[ARM=skill] status={STATUS} reward={REWARD} exception={EXCEPTION_TYPE}
---INJECTED SKILL.md---
{INJECTED_SKILL}

---codex.txt (head + tail)---
{SKILL_CODEX_EXCERPT}

Output STRICT JSON only (no markdown fences). Schema:

{
  "per_arm": {
    "raw":      {"mode": "<mode name or null if MISSING>", "status": "...", "evidence_quote": "<=160 chars verbatim or empty>"},
    "workflow": {...},
    "skill":    {...}
  },
  "deltas": {
    "workflow_vs_raw": {
      "what_changed": "1-2 sentences",
      "net_effect": "fixed" | "regressed" | "unchanged" | "mixed" | "not_comparable",
      "fixed_mode": [],
      "introduced_mode": []
    },
    "skill_vs_raw":       {...same shape...},
    "skill_vs_workflow":  {...same shape...}
  },
  "skill_mechanism": "knowledge_injection | procedural_anchor | failure_warning | none | counterproductive",
  "skill_mechanism_reason": "one sentence",
  "workflow_mechanism": "knowledge_injection | procedural_anchor | failure_warning | none | counterproductive",
  "workflow_mechanism_reason": "one sentence",
  "confidence": "high | medium | low"
}

Rules:
- Use exact mode names from the taxonomy (snake_case). If a trial does not fit any, use "other".
- For MISSING arms, set per_arm.<arm> = null and deltas involving that arm to net_effect="not_comparable".
- fixed_mode / introduced_mode are mode names that the comparison arm eliminated or newly caused respectively (relative to the baseline arm).
- evidence_quote must be verbatim from codex.txt / instruction.md / SKILL.md.
- skill_mechanism = "none" if skill content was not used by agent at all.
- skill_mechanism = "counterproductive" if skill made the run worse than raw.
\end{promptlistingpurple}

\section{Complete Results for Skill Retrieval and Outcome Annotation Ablation}
\label{app:skill-retrieval-results}

\begin{table}[h]
\centering
\scriptsize
\setlength{\tabcolsep}{3.2pt}
\renewcommand{\arraystretch}{1.06}
\resizebox{0.6\columnwidth}{!}{%
\begin{tabular}{ll|c|ccc|ccc}
\toprule
\multirow{2}{*}{\textbf{Pool}} & \multirow{2}{*}{\textbf{$k$}} & \textbf{Top-1} & \multicolumn{3}{c|}{\textbf{Top-3}} & \multicolumn{3}{c}{\textbf{Top-5}} \\
 & & P & P & R & F1 & P & R & F1 \\
\midrule
Random & 5 & 97.7 & 33.0 & 98.9 & 49.4 & -- & -- & -- \\
 & 10 & 95.5 & 32.6 & 97.7 & 48.9 & 19.5 & 97.7 & 32.6 \\
 & 20 & 95.5 & 32.2 & 96.6 & 48.3 & 19.5 & 97.7 & 32.6 \\
 & 50 & 92.0 & 32.2 & 96.6 & 48.3 & 19.5 & 97.7 & 32.6 \\
 & 100 & 84.1 & 30.7 & 92.0 & 46.0 & 19.1 & 95.5 & 31.8 \\
\midrule
Similar & 5 & 70.5 & 31.8 & 95.5 & 47.7 & -- & -- & -- \\
 & 10 & 63.6 & 29.2 & 87.5 & 43.8 & 19.3 & 96.6 & 32.2 \\
 & 20 & 60.2 & 26.9 & 80.7 & 40.3 & 17.5 & 87.5 & 29.2 \\
 & 50 & 56.8 & 24.2 & 72.7 & 36.4 & 16.4 & 81.8 & 27.3 \\
 & 100 & 53.4 & 22.7 & 68.2 & 34.1 & 15.7 & 78.4 & 26.1 \\
\midrule
Dissimilar & 5 & 96.6 & 33.0 & 98.9 & 49.4 & -- & -- & -- \\
 & 10 & 96.6 & 32.2 & 96.6 & 48.3 & 19.8 & 98.9 & 33.0 \\
 & 20 & 96.6 & 32.2 & 96.6 & 48.3 & 19.5 & 97.7 & 32.6 \\
 & 50 & 94.3 & 32.2 & 96.6 & 48.3 & 19.5 & 97.7 & 32.6 \\
 & 100 & 93.2 & 32.2 & 96.6 & 48.3 & 19.3 & 96.6 & 32.2 \\
\bottomrule
\end{tabular}}
\caption{\textbf{Complete Arm 1 embedding-retrieval results on SkillsBench.} Entries report ranking metrics from Qwen3-Embedding-0.6B using task--skill-description similarity. Top-5 is omitted for $k=5$ because it covers the full candidate pool.}
\label{tab:retrieval-full-arm1}
\end{table}

\begin{table}[t]
\centering
\scriptsize
\setlength{\tabcolsep}{3.2pt}
\renewcommand{\arraystretch}{1.04}
\resizebox{0.7\textwidth}{!}{%
\begin{tabular}{ccc|AAA|BBBB}
\toprule
\multirow{2}{*}{\shortstack[c]{\textbf{Agent / Model}}} & \multirow{2}{*}{\shortstack[c]{\textbf{Pool}}} & \multirow{2}{*}{\shortstack[c]{\textbf{$k$}}} & \multicolumn{3}{>{\columncolor{ArmTwoBg}}c|}{\textbf{Arm 2: Agent Selection}} & \multicolumn{4}{>{\columncolor{ArmThreeBg}}c}{\textbf{Arm 3: Real Execution}} \\
\cmidrule(lr){4-6}\cmidrule(l){7-10}
\rowcolor{HeaderGray}
& & & \textbf{P} & \textbf{R} & \textbf{F1} & \textbf{P} & \textbf{R} & \textbf{F1} & \textbf{Succ.} \\
\midrule
\multirow{15}{*}{\shortstack[c]{Gemini CLI\\Gemini-3.1-Pro-Preview}} & Random & 5 & \armtwo{74.4} & \armtwo{75.0} & \armtwo{74.7} & \armthree{18.6} & \armthree{69.4} & \armthree{29.3} & \armthree{39.2} \\
 & Random & 10 & \armtwo{72.7} & \armtwo{72.7} & \armtwo{72.7} & \armthree{7.1} & \armthree{66.4} & \armthree{12.9} & \armthree{38.1} \\
 & Random & 20 & \armtwo{80.1} & \armtwo{81.8} & \armtwo{81.0} & \armthree{3.6} & \armthree{66.0} & \armthree{6.8} & \armthree{39.2} \\
 & Random & 50 & \armtwo{81.2} & \armtwo{84.1} & \armtwo{82.6} & \armthree{1.4} & \armthree{65.8} & \armthree{2.8} & \armthree{37.6} \\
 & Random & 100 & \armtwo{77.4} & \armtwo{83.0} & \armtwo{80.1} & \armthree{0.7} & \armthree{66.0} & \armthree{1.4} & \armthree{39.0} \\
 & Similar & 5 & \armtwo{54.3} & \armtwo{69.3} & \armtwo{60.9} & \armthree{17.6} & \armthree{70.1} & \armthree{28.1} & \armthree{38.8} \\
 & Similar & 10 & \armtwo{61.3} & \armtwo{79.5} & \armtwo{69.2} & \armthree{7.0} & \armthree{67.3} & \armthree{12.6} & \armthree{38.5} \\
 & Similar & 20 & \armtwo{58.4} & \armtwo{77.3} & \armtwo{66.6} & \armthree{3.7} & \armthree{66.2} & \armthree{6.9} & \armthree{34.0} \\
 & Similar & 50 & \armtwo{59.3} & \armtwo{76.1} & \armtwo{66.7} & \armthree{1.4} & \armthree{66.2} & \armthree{2.7} & \armthree{36.7} \\
 & Similar & 100 & \armtwo{55.4} & \armtwo{73.9} & \armtwo{63.3} & \armthree{0.7} & \armthree{66.7} & \armthree{1.4} & \armthree{36.1} \\
 & Dissimilar & 5 & \armtwo{74.4} & \armtwo{76.1} & \armtwo{75.3} & \armthree{14.6} & \armthree{63.2} & \armthree{23.7} & \armthree{34.0} \\
 & Dissimilar & 10 & \armtwo{80.1} & \armtwo{81.8} & \armtwo{81.0} & \armthree{8.7} & \armthree{66.0} & \armthree{15.3} & \armthree{38.8} \\
 & Dissimilar & 20 & \armtwo{82.4} & \armtwo{83.0} & \armtwo{82.7} & \armthree{5.1} & \armthree{65.5} & \armthree{9.5} & \armthree{35.6} \\
 & Dissimilar & 50 & \armtwo{79.3} & \armtwo{80.7} & \armtwo{80.0} & \armthree{1.6} & \armthree{64.4} & \armthree{3.1} & \armthree{38.1} \\
 & Dissimilar & 100 & \armtwo{83.5} & \armtwo{84.1} & \armtwo{83.8} & \armthree{0.7} & \armthree{61.6} & \armthree{1.5} & \armthree{34.5} \\
\midrule
\multirow{15}{*}{\shortstack[c]{Codex\\GPT-5.4}} & Random & 5 & \armtwo{81.8} & \armtwo{84.1} & \armtwo{82.9} & \armthree{33.1} & \armthree{41.6} & \armthree{36.8} & \armthree{24.3} \\
 & Random & 10 & \armtwo{83.0} & \armtwo{86.4} & \armtwo{84.6} & \armthree{39.2} & \armthree{59.7} & \armthree{47.3} & \armthree{35.5} \\
 & Random & 20 & \armtwo{84.1} & \armtwo{87.5} & \armtwo{85.8} & \armthree{35.3} & \armthree{69.3} & \armthree{46.8} & \armthree{40.9} \\
 & Random & 50 & \armtwo{71.7} & \armtwo{87.5} & \armtwo{78.8} & \armthree{15.8} & \armthree{63.5} & \armthree{25.4} & \armthree{35.0} \\
 & Random & 100 & \armtwo{62.2} & \armtwo{85.2} & \armtwo{71.9} & \armthree{8.1} & \armthree{73.6} & \armthree{14.6} & \armthree{44.8} \\
 & Similar & 5 & \armtwo{51.9} & \armtwo{81.8} & \armtwo{63.5} & \armthree{51.3} & \armthree{72.4} & \armthree{60.0} & \armthree{44.5} \\
 & Similar & 10 & \armtwo{44.4} & \armtwo{83.0} & \armtwo{57.8} & \armthree{37.5} & \armthree{66.0} & \armthree{47.9} & \armthree{40.7} \\
 & Similar & 20 & \armtwo{35.8} & \armtwo{77.3} & \armtwo{48.9} & \armthree{27.7} & \armthree{66.1} & \armthree{39.1} & \armthree{44.3} \\
 & Similar & 50 & \armtwo{37.8} & \armtwo{76.1} & \armtwo{50.5} & \armthree{13.1} & \armthree{61.8} & \armthree{21.6} & \armthree{42.3} \\
 & Similar & 100 & \armtwo{31.9} & \armtwo{70.5} & \armtwo{43.9} & \armthree{6.7} & \armthree{54.3} & \armthree{11.9} & \armthree{43.0} \\
 & Dissimilar & 5 & \armtwo{83.3} & \armtwo{85.2} & \armtwo{84.3} & \armthree{42.5} & \armthree{63.0} & \armthree{50.7} & \armthree{37.3} \\
 & Dissimilar & 10 & \armtwo{83.0} & \armtwo{85.2} & \armtwo{84.1} & \armthree{29.7} & \armthree{61.3} & \armthree{40.0} & \armthree{35.0} \\
 & Dissimilar & 20 & \armtwo{82.8} & \armtwo{85.2} & \armtwo{84.0} & \armthree{12.7} & \armthree{60.5} & \armthree{21.0} & \armthree{31.8} \\
 & Dissimilar & 50 & \armtwo{77.5} & \armtwo{85.2} & \armtwo{81.2} & \armthree{7.2} & \armthree{65.1} & \armthree{12.9} & \armthree{39.5} \\
 & Dissimilar & 100 & \armtwo{72.0} & \armtwo{85.2} & \armtwo{78.0} & \armthree{2.7} & \armthree{60.2} & \armthree{5.2} & \armthree{38.2} \\
\bottomrule
\end{tabular}}
\caption{\textbf{Complete Arm 2 and Arm 3 retrieval results on SkillsBench.} Rows correspond to agent--model, distractor regime, and pool size. Arm 2 reports explicit-selection precision, recall, and F1; Arm 3 reports parsed actual-use precision, recall, F1, and downstream success. Dashes indicate excluded entries.}
\label{tab:retrieval-full-arm2-arm3}
\end{table}

\begin{table*}[t]
\centering
\scriptsize
\setlength{\tabcolsep}{3.4pt}
\renewcommand{\arraystretch}{1.08}
\resizebox{0.7\textwidth}{!}{%
\begin{tabular}{lll|*{6}{c}}
\toprule
\textbf{Agent + Model} & \textbf{Benchmark} & \textbf{Creator} & \textbf{5s0f} & \textbf{4s1f} & \textbf{3s2f} & \textbf{2s3f} & \textbf{1s4f} & \textbf{0s5f} \\
\midrule
\multirow{6}{*}{\shortstack[l]{Codex\\GPT-5.3-Codex}}
& \multirow{2}{*}{TB2} & normal  & 0.7548 & 0.7290 & 0.7806 & 0.6839 & 0.7097 & 0.5161 \\
& & no-hint & 0.7677 & 0.7355 & 0.5871 & 0.4968 & 0.5548 & 0.3871 \\
\cmidrule(l){2-9}
& \multirow{2}{*}{SB} & normal  & 0.7250 & 0.6167 & 0.6250 & 0.7083 & 0.6167 & 0.4500 \\
& & no-hint & 0.6667 & 0.6417 & 0.5583 & 0.5000 & 0.5083 & 0.3500 \\
\cmidrule(l){2-9}
& \multirow{2}{*}{TB-Pro} & normal  & 0.7455 & 0.7939 & 0.7333 & 0.6667 & 0.5818 & 0.4303 \\
& & no-hint & 0.8364 & 0.6606 & 0.5758 & 0.5152 & 0.4848 & 0.3758 \\
\midrule
\multirow{6}{*}{\shortstack[l]{Gemini CLI\\Gemini-3.1-Pro-Preview}}
& \multirow{2}{*}{TB2} & normal  & 0.7923 & 0.7615 & 0.7462 & 0.7000 & 0.6923 & 0.4769 \\
& & no-hint & 0.4231 & 0.4923 & 0.4000 & 0.3692 & 0.5231 & 0.4308 \\
\cmidrule(l){2-9}
& \multirow{2}{*}{SB} & normal  & 0.7429 & 0.6190 & 0.6667 & 0.6762 & 0.6000 & 0.4095 \\
& & no-hint & 0.6190 & 0.5143 & 0.4095 & 0.4190 & 0.4095 & 0.4000 \\
\cmidrule(l){2-9}
& \multirow{2}{*}{TB-Pro} & normal  & 0.6692 & 0.6308 & 0.5462 & 0.5077 & 0.5692 & 0.4615 \\
& & no-hint & 0.6154 & 0.5769 & 0.6385 & 0.5923 & 0.5692 & 0.5154 \\
\bottomrule
\end{tabular}}
\caption{\textbf{Complete numerical results for the outcome-annotation ablation.} Entries report downstream success for skills constructed under the indicated trajectory mixture. \textit{normal} exposes source-trajectory outcomes during construction; \textit{no-hint} withholds them. Terminal-Bench-Pro entries use 130 trials per condition, with missing or infrastructure-error trials counted as failures.}
\label{tab:nohint-skills-singlecol}
\end{table*}

\end{document}